\documentclass{article}

\usepackage[numbers]{natbib}

\usepackage[preprint]{neurips_2023}

\usepackage[utf8]{inputenc} 
\usepackage[T1]{fontenc}    
\usepackage{url}            
\usepackage{booktabs}       
\usepackage{amsfonts}       
\usepackage{nicefrac}       
\usepackage{microtype}      
\usepackage{epsfig}
\usepackage{amsmath}
\usepackage{mathtools}
\usepackage{caption}
\usepackage{subcaption}
\usepackage{algorithm}
\usepackage{algpseudocode}
\usepackage{authblk}
\usepackage{bbm}
\usepackage{pifont}
\usepackage{color}

\usepackage{amsmath,amsfonts,bm}

\def\eqref#1{equation~\ref{#1}}

\def\1{\bm{1}}

\def\eps{{\epsilon}}

\DeclareMathAlphabet{\mathsfit}{\encodingdefault}{\sfdefault}{m}{sl}
\SetMathAlphabet{\mathsfit}{bold}{\encodingdefault}{\sfdefault}{bx}{n}

\DeclareMathOperator*{\argmin}{arg\,min}

\usepackage{multirow}
\usepackage{textcomp}
\usepackage{comment}
\usepackage{graphicx,wrapfig,lipsum}
\usepackage{diagbox}
\usepackage{amsthm}

\usepackage[x11names]{xcolor}
\usepackage{colortbl}
\usepackage[
  colorlinks,
]{hyperref}

\usepackage{cleveref}
\usepackage{mwe} 

\AtBeginDocument{\hypersetup{citecolor=DodgerBlue3}}

\usepackage{booktabs,arydshln}

\makeatletter
\def\adl@drawiv#1#2#3{%
        \hskip.5\tabcolsep
        \xleaders#3{#2.5\@tempdimb #1{1}#2.5\@tempdimb}%
                #2\z@ plus1fil minus1fil\relax
        \hskip.5\tabcolsep}
\newcommand{\cdashlinelr}[1]{%
  \noalign{\vskip\aboverulesep
           \global\let\@dashdrawstore\adl@draw
           \global\let\adl@draw\adl@drawiv}
  \cdashline{#1}
  \noalign{\global\let\adl@draw\@dashdrawstore
           \vskip\belowrulesep}}
\makeatother

\newcommand{\etal}{\textit{et al.~}}

\title{Fill-Up: Balancing Long-Tailed Data \\ with Generative Models}
\author{%
\textbf{Joonghyuk Shin} \hspace{1cm} \textbf{Minguk Kang} \hspace{1cm} \textbf{Jaesik Park}\\
Pohang University of Science and Technology~(POSTECH), South Korea\\
\texttt{\{jhshin4727,mgkang,jaesik.park\}@postech.ac.kr} \\
}
 
\begin{document}
\maketitle
\begin{abstract}
Modern text-to-image synthesis models have achieved an exceptional level of photorealism, generating high-quality images from arbitrary text descriptions. In light of the impressive synthesis ability, several studies have exhibited promising results in exploiting generated data for image recognition. However, directly supplementing data-hungry situations in the real-world (\textit{e.g.} few-shot or long-tailed scenarios) with existing approaches result in marginal performance gains, as they suffer to thoroughly reflect the distribution of the real data. Through extensive experiments, this paper proposes a new image synthesis pipeline for long-tailed situations using Textual Inversion. The study demonstrates that generated images from textual-inverted text tokens effectively aligns with the real domain, significantly enhancing the recognition ability of a standard ResNet50 backbone. We also show that real-world data imbalance scenarios can be successfully mitigated by filling up the imbalanced data with synthetic images. In conjunction with techniques in the area of long-tailed recognition, our method achieves state-of-the-art results on standard long-tailed benchmarks when trained from scratch.
\end{abstract}

\section{Introduction}
The intricate landscape of real-world data often reveals an inherent imbalance, where the abundance of common samples dominate the scarce presence of their rare counterparts~\cite{lin2014microsoft, liu2015deep, van2018inaturalist, liu2019large, johnson2019survey, hedderich2021survey, yang2022survey, henning2022survey}. Long-tailed recognition is an area that aims to acquire balanced knowledge in this lopsided situation, irrespective of different sample sizes for each class. Several works tried to mitigate this class imbalance challenge through diverse approaches such as class-balanced sampling~\cite{chawla2002smote, drummond2003c4, han2005borderline, van2007experimental}, loss re-weighting~\cite{huang2016learning, khan2017cost, cui2019class, cao2019learning, ren2020balanced} and data augmentation~\cite{chawla2002smote, han2005borderline, he2008adasyn, zhang2018mixup, yun2019cutmix, cubuk2020randaugment, chou2020remix, kim2020m2m, park2021cmo, zang2021fasa}. Recent approaches also include decoupling methods~\cite{Kang2020Decoupling, zhang2021bag, zhong2021improving}, and the integration of mixture-of-experts techniques~\cite{wang2021longtailed, li2022nested, aimar2022balanced, jin2023long}. 

Given that the underlying motivation of these approaches revolves around re-balancing data distributions, an intuitive approach would be to fill up the imbalanced data using synthetic samples. However, training traditional class conditional generative models under severe data imbalance is known to be highly challenging~\cite{zhao2020diffaugment,Karras2020TrainingGA,shahbazi2022collapse, rangwani2022improving, rangwani2023noisytwins}, hindering the practical usages of well-established generative models such as VAE~\cite{kingma2013auto} and GAN~\cite{Goodfellow2014GenerativeAN}.

The rapid development of generative models~\cite{ho2020denoising, nichol2022glide, ramesh2021zero, rombach2022high, ramesh2022hierarchical, saharia2022photorealistic, balaji2022ediffi, kang2023gigagan}, has ushered in advancements across diverse areas of machine learning. Synthetic data generation is one area where these photorealistic images with delicate details exhibit promising results. Multiple studies~\cite{he2023is, sariyildiz2023fake, shipard2023diversity, bansal2023leaving} have proposed efficient methods for zero/few-shot classification, elucidating the robustness of features acquired from synthetic data and their efficacy in transfer learning scenarios. 
However, these zero-shot methods, predominantly reliant on prompting techniques, often exhibit marginal performance gains in the presence of real data, constraining their practical usages in real-world long-tailed scenarios. 

In this paper, with the help of recent large-scale text-to-image synthesis models, we effectively address this issue by adopting per-class optimization of randomly initialized text tokens. 
This work starts by evaluating a range of novel and established strategies devised for synthetic image generation. After a thorough examination, we empirically discover that our new approach based on Textual Inversion~\cite{gal2022image} exhibits a remarkable ability to generate diverse images that align with the real domain without using class-related text information. Notably, by exclusively leveraging information within the image domain, our method showcases its capability to generate classes that were underrepresented by the original text-conditioned pretrained model. In conjunction with the simple yet effective Balanced Softmax loss~\cite{ren2020balanced} and the proposed two-stage training procedure, our method attains state-of-the-art results on standard long-tailed classification benchmarks~\cite{liu2019large, van2018inaturalist} when trained from scratch. Our method especially demonstrates significantly higher accuracies in few-shot scenarios involving classes with fewer than 20 samples.

Overall, our work makes the following key contributions: (1) We offer a comprehensive evaluation of various generation strategies based on text-to-image synthesis models, incorporating diverse ablations and providing valuable insights. (2) We propose a novel image generation method tailored for real-world's data imbalance scenarios involving Textual Inversion, which outperforms existing generation methods. (3) In combination with our suggested two-stage training procedure and the Balanced Softmax loss, our method achieves state-of-the-art results on standard long-tailed benchmarks when trained from scratch.
\section{Related work}

\subsection{Large scale text-to-image synthesis models}
\label{sec:Background_LargeScale}
Recent advances in diffusion models~\cite{ho2020denoising, dhariwal2021diffusion, nichol2021improved, song2021denoising} and multimodal learning~\cite{radford2021learning, jia2021scaling} have led to significant progress in the area of text-conditioned image synthesis, which was considered to be extremely challenging with traditional conditional GANs~\cite{Goodfellow2014GenerativeAN, Brock2019LargeSG, karras2019style, sauer2022stylegan, kang2022studiogan}. Some notable examples of diffusion models include GLIDE~\cite{nichol2022glide}, Dall$\cdot$E 2~\cite{ramesh2022hierarchical}, Imagen~\cite{saharia2022photorealistic}, Stable Diffusion~\cite{rombach2022high}, and eDiff-I~\cite{balaji2022ediffi}. Autoregressive and GAN-based models such as Dall$\cdot$E~\cite{ramesh2021zero}, Parti~\cite{yu2022scaling}, GigaGAN~\cite{kang2023gigagan}, and StyleGAN-T~\cite{sauer2023stylegan} have also shown very promising results. In conjunction with the advancements of generative models, there has been extensive research on methods to personalize synthesized outputs. Representative works include Textual Inversion~\cite{gal2022image}, DreamBooth~\cite{ruiz2022dreambooth}, and Custom Diffusion~\cite{kumari2022multi}, each focusing on fine-tuning different parts of diffusion models. 

Among several options, we utilize Stable Diffusion, a text-conditioned variant of the Latent Diffusion Model~\cite{rombach2022high} (LDM), trained on LAION-5B~\cite{schuhmann2022laion}. LDM-based models leverage pretrained encoder and decoder networks to perform forward and reverse diffusion processes in the latent space, reducing the expensive compute overhead of the diffusion family. In this paper, we utilize Stable Diffusion v1.5~\cite{sdv1-5}, unless stated otherwise.

\subsection{Learning from synthetic data}
\label{sec:Background_SynthGen}
Building real-world datasets is often challenging due to the difficulty of collecting a balanced set of images, as well as the inherent privacy and bias concerns involved. Simulation pipelines from 3D engines~\cite{richter2016playing, peng2017visda} or pretrained class-conditional GANs have been utilized as data sources. For instance, GANs can generate pixel-wise annotations for semantic segmentation~\cite{zhang2021datasetgan, li2022bigdatasetgan} and improve representation learning through different view generation~\cite{chai2021ensembling, jahaniangenerative}. Nonetheless, synthetic data generated from 3D engines may have domain gaps with real data and show limited diversity. Additionally, training generative models from scratch usually requires a large amount of well-curated data, further restricting the applicability of existing methods in data-scarce situations. 

More recent approaches~\cite{he2023is, sariyildiz2023fake, shipard2023diversity, azizi2023synthetic, bansal2023leaving} incorporate general-purpose text-conditioned diffusion models. He~\etal~\cite{he2023is} employed GLIDE to generate synthetic images and used them to fine-tune CLIP~\cite{radford2021learning}. Sariyildiz~\etal~\cite{sariyildiz2023fake} focused more on ImageNet~\cite{Deng2009ImageNetAL} and trained models from scratch using various prompting methods. Their common findings suggest that features learned from the synthetic data can be robustly transferred to downstream tasks. These results highlight the crucial roles of classifier-free guidance scale and the number of images in achieving high diversity and classification accuracy. Different from the above approaches, Li~\etal~\cite{li2023your} directly turned pretrained diffusion models into zero-shot classifiers, and Azizi~\etal~\cite{azizi2023synthetic} fine-tuned Imagen~\cite{saharia2022photorealistic} with ImageNet~\cite{Deng2009ImageNetAL} to produce class-conditional model that can potentially improve ImageNet classification. The use of synthetic data from diffusion models is also becoming prevalent in other computer vision domains, such as semantic segmentation~\cite{wu2023diffumask}, object detection~\cite{lin2023explore}, and semi-supervised learning~\cite{you2023diffusion}. 

However, the synthetic images generated through the aforementioned methods, which target zero-shot scenarios, present an issue of distribution misalignment~\cite{he2023is}. Consequently, the direct application of these zero-shot methods in the presence of real data (\textit{e.g.} long-tailed or few-shot) leads to the generation of noisy samples that deviate from the distribution of the real data. As a result, despite the availability of valuable real samples, these approaches struggle to further improve the performance, as discussed in Sec.~\ref{sec:Experiments_Efficiency}.

Considering the challenge of learning from imbalanced data for generative models~\cite{shahbazi2022collapse, rangwani2022improving, rangwani2023noisytwins}, simple strategies such as training or fine-tuning the entire generative model under imbalanced data, are also not practical. Moreover, employing a na\"ive per-class fine-tuning approach would result in considerable inefficiency, as it necessitates a substantial amount of model weights to be saved with an increasing number of classes. In this study, we effectively circumvent these issues by adopting an efficient per-class tuning strategy for text tokens.

\subsection{Long-tailed recognition}
\label{sec:Background_LT}

\paragraph{Re-sampling and re-weighting.} 
The classical approach for handling long-tailed data involves re-sampling and re-weighting. Re-sampling methods aim to generate class-balanced data by over-sampling of minority classes~\cite{van2007experimental, buda2018systematic} and under-sampling of majority classes~\cite{van2007experimental, mani2003knn}. However, it is known that over-sampling can lead to overfitting on tail classes~\cite{sarafianos2018deep}, while under-sampling may result in losing valuable information from head classes~\cite{park2021cmo}. Re-weighting methods tackle this issue by assigning different weights on either the class level~\cite{khan2017cost, cui2018large, ren2018learning, cao2019learning, khan2019striking} or the instance level~\cite{lin2017focal, shu2019meta}, resulting in distinct losses for each class. Nevertheless, it is known that re-weighting can introduce instability during the training, particularly when dealing with the extreme class imbalance in large-scale datasets. 

\paragraph{Data augmentation.} 
Augmenting real samples~\cite{zhang2018mixup, yun2019cutmix, cubuk2020randaugment} or introducing synthetic samples~\cite{kim2020m2m, mullick2019generative, kozerawski2020blt} is also promising direction. Modern augmentation strategies involve transferring information from the majority classes to the minority classes~\cite{park2021cmo}. In the domain of synthetic sample generation, some studies~\cite{chawla2002smote, zang2021fasa} have shown successful results by generating samples at the feature level. Closer to our work, GAMO~\cite{mullick2019generative} adopts a three-player adversarial game to generate feature-level samples and uses VAE~\cite{kingma2013auto} to generate image-level samples. However, given the easy-to-collapse nature of the adversarial game and the difficulty of learning imbalanced distributions, GAMO struggles to generate high-quality images, resulting in a marginal performance gain that is further constrained to smaller datasets. To the best of our knowledge, our approach is the first to generate realistic image-level samples that can actually perform on par or even surpass other state-of-the-art long-tailed methods.

\paragraph{Other long-tailed methods.}
Recently, decoupling the learning of representation and classifier, following ~\cite{Kang2020Decoupling}, has become one of the mainstream approaches~\cite{zhang2021bag, zhong2021improving}. Another advantage of the decoupling method is that it can be readily combined with other techniques, such as re-sampling or re-weighting, further amplifying its potential. Meta-learning~\cite{ren2020balanced, jamal2020rethinking, wang2020meta, li2021metasaug}, contrastive learning~\cite{cui2021parametric, suh2023long}, and ensembling~\cite{wang2021longtailed, li2022nested, aimar2022balanced, jin2023long} based methods are additional sources of recent successes. The most recent approaches~\cite{ma2021simple, tian2022vl, dong2023lpt} involve using a powerful multimodal model, CLIP~\cite{radford2021learning}. By incorporating additional text data acquired from the web (\textit{e.g., Wikipedia})~\cite{tian2022vl} or additional prompt-tuning~\cite{dong2023lpt}, recent works further improved the performance of the pretrained CLIP. However, these methods heavily rely on the strong zero-shot performance of CLIP, which already surpasses the performances of most state-of-the-art methods. In this work, we mainly focus on learning imbalanced distribution from scratch, with a specific emphasis on acquiring robust representation exclusively in the domain of images. 

\section{Filling up long-tailed data with generative models}

\subsection{Synthetic image generation strategies}
\label{sec:Method_strategies}
\begin{figure}[t!]
    \centering
    \begin{subfigure}[t]{0.99\textwidth} 
    \includegraphics[width=1.0\linewidth]{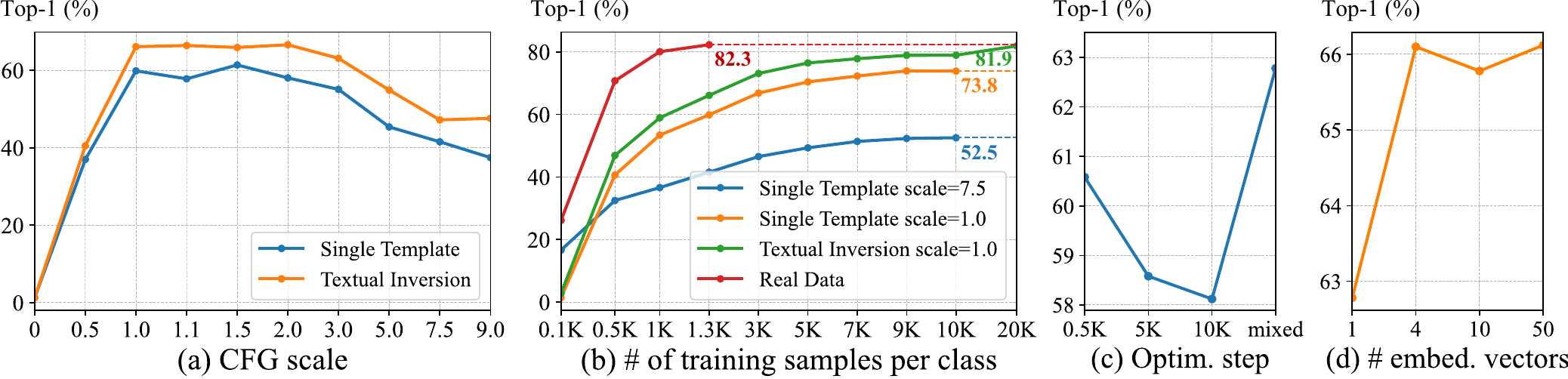}
    \label{fig:method_ablation}
    \end{subfigure}
    \begin{subfigure}[t]{0\textwidth} 
         \includegraphics[width=\textwidth]{example-image-a}
         \phantomcaption
         \label{fig:imga}   
    \end{subfigure}
    \begin{subfigure}[t]{0\textwidth} 
         \includegraphics[width=\textwidth]{example-image-b}
         \phantomcaption
         \label{fig:imgb}   
    \end{subfigure}
    \begin{subfigure}[t]{0\textwidth} 
         \includegraphics[width=\textwidth]{example-image-c}
         \phantomcaption
         \label{fig:imgc}   
    \end{subfigure}
    \begin{subfigure}[t]{0\textwidth} 
         \includegraphics[width=\textwidth]{example-image-c}
         \phantomcaption
         \label{fig:imgd}   
    \end{subfigure}
    \vspace{-8mm}
    \caption{(a) Impact of classifier-free guidance scale on top-1 and top-5 accuracy. (b) Improvements in top-1 accuracy when scaling number of training samples per class for different generation strategies. (c, d) Performance of \textit{Textual Inversion} based method with varying optimization steps used for generation and increasing size of optimizable text embedding.}
    \vspace{-3mm}
\end{figure}

Our study begins by evaluating diverse synthetic data generation approaches and selecting the best fit for our data-scarce scenario. As pointed out by previous work~\cite{sariyildiz2023fake, shipard2023diversity, azizi2023synthetic}, the problem of diversity has been problematic in utilizing text-to-image synthesis models for image recognition. Since cutting-edge text-to-image synthesis models prioritize aligning the visual appearance with input text prompts, they tend to generate high-fidelity but low-diversity images. A straightforward approach to balance fidelity and diversity is adjusting the classifier-free guidance scale~\cite{ho2021classifierfree} as shown in Fig.~\ref{fig:imga}.

Recent models that employ classifier-free guidance, such as Stable Diffusion, provide the capability to generate more diverse but slightly noisier samples by decreasing the default guidance scale $w$, which is set to 7.5 in the case of Stable Diffusion. Eq.~(\ref{eq:cfg}) demonstrates the process of classifier-free guidance in LDM variants~\cite{rombach2022high}, where $\epsilon$ represents a denoising U-Net, $t$ denotes a timestamp, $z$ refers to the latent variable, and $c$ is a conditioning vector. Higher guidance enforces stronger conditioning of input prompts, and it is worth noting that classifier-free guidance is no longer adopted when the scale is set to 1.0, reducing the necessity of an additional reverse process for null-conditioning. 
\begin{align}
    \tilde{\epsilon}_{t} = \epsilon_{\theta}(z_{t}) + w(\epsilon_{\theta}(z_{t}, c) - \epsilon_{\theta}(z_{t})). 
    \label{eq:cfg}
\end{align}
To effectively address the issue of diversity, we thoroughly evaluate a range of novel and established methods designed to generate feature-rich images and extensively compare their performances at different guidance scales. For evaluation of the mentioned methods, we use IN100 and IN100-LT from Sec.~\ref{sec:Experiments_Datasets}, generating 1,300 images per class using DDIM sampler~\cite{song2020denoising} with 50 diffusion steps. We present comprehensive qualitative results in Appendix.

\begin{figure}[t!]
    \centering
    \includegraphics[width=0.96\linewidth]{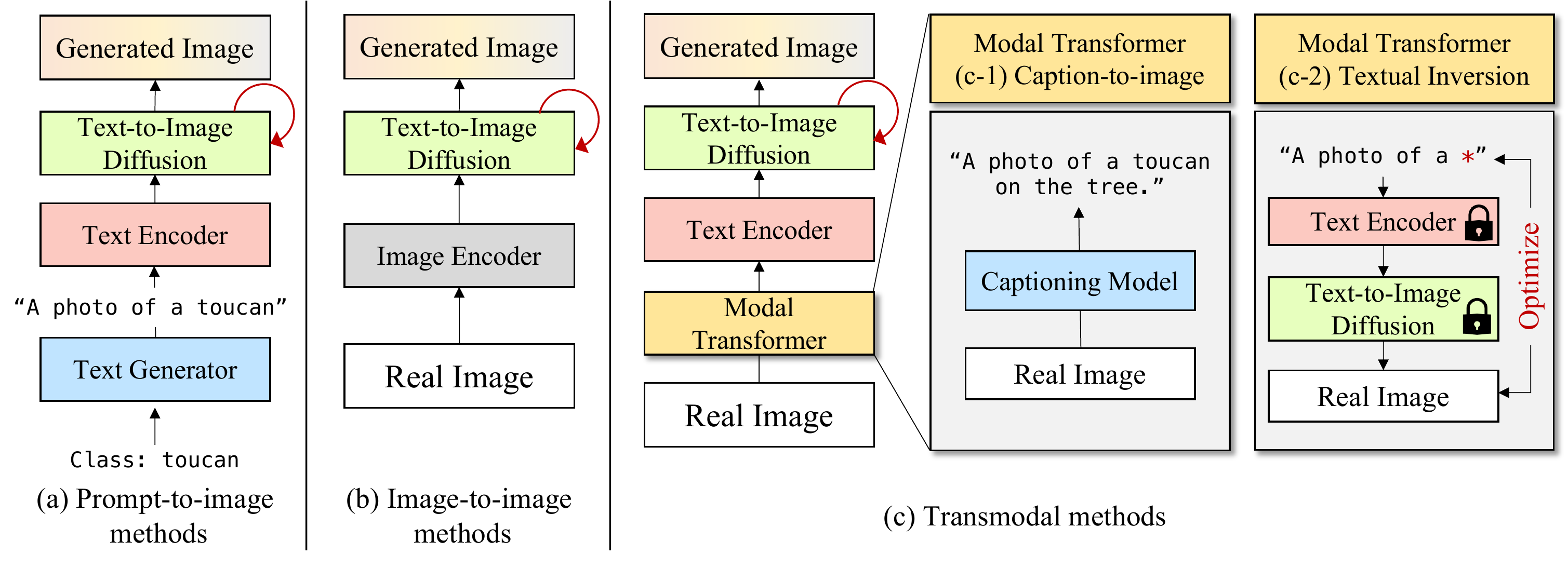}
    \caption{Illustration of our different generation strategies: prompt-to-image methods (\textit{Single template, CLIP template, T5~\cite{raffel2020exploring}, and Flan-T5 XXL~\cite{chung2022scaling}}), image-to-image methods (\textit{Image variation~\cite{sd1-variation} and Reimagine~\cite{sd2-reimagine}}), and transmodal methods (\textit{Captioning~\cite{li2023blip} and Textual Inversion~\cite{gal2022image}}).}
    \label{fig:Figure_cGANs}
    \vspace{-3mm}
\end{figure}

\paragraph{Prompt-to-image based methods.}
\label{sec:Method_p2i} The most straightforward method to utilize text-to-image synthesis models is to use na\"ive prompts (\textit{e.g.} "a photo of a \{\texttt{CLS}\}"). However, this method, namely, \textit{single template}, is known to produce images that substantially lack diversity. He~\etal~\cite{he2023is} and Yuan~\etal~\cite{yuan2022not} introduce the use of large-scale language models and CLIP templates to alleviate this issue. He~\etal use T5~\cite{raffel2020exploring} fine-tuned on CommonGen~\cite{lin-etal-2020-commongen} combined with CLIP filtering to remove noisy generated sentences. Building upon previous work, we experiment with the same T5~\cite{t5-huggingface}, a larger language model (Flan-T5 XXL~\cite{chung2022scaling} fine-tuned in Alpaca style~\cite{alpaca, flan-t5-huggingface}) and a community-famous prompt-extender~\cite{prompt-extend-huggingface} for Stable Diffusion which is a GPT-2 model fine-tuned on prompts from DiffusionDB~\cite{wang2022diffusiondb}. Table~\ref{method_performance} presents the results for four settings: \textit{single template, CLIP templates, T5, and Flan-T5 XXL}. Additional details (\textit{e.g.} decoding strategies) and the performance for \textit{Prompt-Extend} can be found in Appendix. We generally observe enhancements with more diverse prompts when the scale is $w=7.5$, but \textit{single template} performs the best as the scale becomes $w=1.0$. With more noise engaged, we believe promoting diversity via complex prompts harms precision by compromising important details of the original class.

\paragraph{Image-to-image based methods.}
\label{sec:Method_i2i}
While the above methods can generally be considered as zero-shot methods, the approaches from here require real samples. For image-to-image based methods, we utilize the state-of-the-art Stable Diffusion Reimagine~\cite{sd2-reimagine}, which replaces the CLIP text encoder with the CLIP image encoder. Since the source images are fully encoded and the reverse process does not originate from the noised source images, this method can produce images that differ in both overall structure and high-frequency details. For a fair comparison, we also test a similar variant called Image Variation~\cite{sd1-variation} based on Stable Diffusion v1.3, as Reimagine is based on Stable Diffusion v2.1. We adopt real samples of IN100-LT described in Sec~\ref{sec:Experiments_Datasets} as source images to evalute the robustness of these methods in real-world data-scarce scenarios. Experimental results of these two settings (\textit{Reimagine and Image Variation}) are elaborated in Table~\ref{method_performance}.
Broadly speaking, these methods perform worse than \textit{single template}, and through qualitative review, we attribute this to specific circumstances where a "reimagined" object replaces the crucial component of an image containing class-specific information. This observation aligns with the fact that this method achieves a relatively high top-5 accuracy considering its performance on top-1. Technical details and qualitative results are provided in Appendix.

\paragraph{Transmodal methods.}
\label{sec:Method_i2p2i}
Here, we propose two novel methods that first generate prompts from real samples, then subsequently synthesize images from these prompts. We refer to this process as a "Transmodal" or simply "image-to-prompt-to-image" approach. The first method utilizes the state-of-the-art captioning model BLIP2~\cite{li2023blip} to generate prompts and uses these prompts for sampling. The second method utilizes Textual Inversion~\cite{gal2022image}. Using few sets of images, we optimize a text token "*" per every class and generate images based on the prompt "a photo of a *". The results of \textit{Captioning and Textual Inversion} can be found in Table~\ref{method_performance}, while additional details are provided in Appendix. We observe that~\textit{Textual Inversion} consistently outperforms all other methods by a notable margin. 

\begin{table}[t!]
\caption{Comparison of accuracies among diverse generation methods for IN100. Each method generated 130K images (1,300 images per class) and synthetic images were exclusively used to train ResNet50~\cite{He_2016_CVPR}. Low guidance indicates a scale of 1.0 and high guidance denotes 7.5, except for \textit{Image Variation}~\cite{sd1-variation} (3.0) and \textit{Reimagine}~\cite{sd2-reimagine} (10.0).}
\label{method_performance}
\centering
\vspace{1mm}
\resizebox{0.99\textwidth}{!}{
\begin{tabular}{lcccccccc}
\cmidrule[1.0pt]{1-9}
    &  \multicolumn{4}{c}{\textbf{Prompt-to-Image}} &  \multicolumn{2}{c}{\textbf{Image-to-Image}} & \multicolumn{2}{c}{\textbf{Transmodal}}\\
    \cmidrule[1.0pt]( r){2-5}
    \cmidrule[1.0pt]( lr){6-7}
    \cmidrule[1.0pt](l ){8-9}
    \textbf{Method} & \text{Single} & \text{CLIP} & \multirow{2}*[0.0ex]{\text{T5}} & \text{Flan-T5}  & \text{Image} & \multirow{2}*[0.0ex]{\text{Reimagine}} & \text{Image} & \text{Textual} \\
    \textbf{(Top-1/Top-5)}& \multirow{2}*[1.5ex]\text{Template} & \multirow{2}*[1.5ex]\text{Templates} &  & \multirow{2}*[1.5ex]\text{-XXL}  & \multirow{2}*[1.5ex]\text{Variation} &  & \multirow{2}*[1.5ex]\text{Captioning} & \multirow{2}*[1.5ex]\text{Inversion} \\
    \cmidrule[1.0pt]{1-9}
Low Guidance    & 59.9 / 83.7   &  57.2 / 82.2      & 45.8 / 71.5    & 51.6 / 77.3    & 47.2 / 77.1    & 54.2 / 80.7    & 55.2 / 83.7    & \textbf{62.8} / \textbf{86.2} \\
High Guidance   & 41.5 / 67.5   &  48.7 / 73.6      & 36.0 / 61.0    & 41.3 / 67.3    & 42.2 / 70.9    & 41.5 / 68.7    & 45.6 / 73.6    & \textbf{49.2} / \textbf{74.3} \\
\cmidrule[1.0pt]{1-9}
\end{tabular}}
\vspace{-3mm}
\end{table}

\paragraph{Textual Inversion details.}
\label{sec:Method_ti}
Textual Inversion~\cite{gal2022image} is an optimization-based personalization method designed to effectively learn the optimal text vector that best explains the concept of given samples. Keeping Stable Diffusion's encoder ($\mathcal{E}$), denoising U-net~($\epsilon_{\theta})$, and CLIP text encoder~($c_{\theta}$) frozen, Textual Inversion aims to optimize the text embedding~($v_{*}$), which is a continuous vector representation of the text token~($y$), using the following objective.
\begin{align}
    v_{*} = \argmin_{v} \mathbb{E}_{z \sim \mathcal{E}(x), y, \epsilon \sim \mathcal{N}(0, 1), t}{\Big[ \Vert \epsilon - \epsilon_{\theta}( z_{t},t, c_{\theta}(y)) \Vert_{2}^{2}\Big]}.
\end{align}
Our empirical results demonstrate that Textual Inversion can efficiently learn class-level concepts given numerous images (\textit{i.e.} head classes) and also effectively capture decisive features given few number of images (\textit{i.e.} tail classes), which is evident considering its initial tailoring for personalization tasks involving 3-5 images. Furthermore, the experiment in Fig.~\ref{fig:imgc} reveals that class-level information is already well-learned in the early stages (500 steps) of optimization, which reduces the need for prolonged training. From this observation, we adopt a simple heuristic based on the number of images that restricts excessive training time and leverage checkpoints from various steps to enhance diversity. Lastly, we observe from Fig.~\ref{fig:imgd} that enlarging the token embedding's capacity from a single vector of 1$\times$768 to 4$\times$768 significantly improves top-1 accuracy, allowing finer details to be captured~\cite{ti-issue}. Yet, further increasing the size of embedding (\textit{e.g.} 10$\times$768) did not yield additional gains, suggesting that a text embedding of shape 4$\times$768 can adequately represent most classes.

Starting from a random text token "*", Textual Inversion consistently proves to be effective for long-tailed data by generating images that align well with the target domain, regardless of the sample size, without relying on any class-related text information. Especially, the high scaling efficiency of this method described in Fig.~\ref{fig:imgb}, provides compelling evidence of the strong potential inherent in textual-inverted tokens. Our qualitative results even suggest its capability to generate classes that were previously underrepresented by conventional text-conditioning. More information on the mentioned heuristic and training, as well as supplementary qualitative results, can be found in Appendix.

\begin{figure}[t!]
    \centering
    \includegraphics[width=0.94\linewidth]{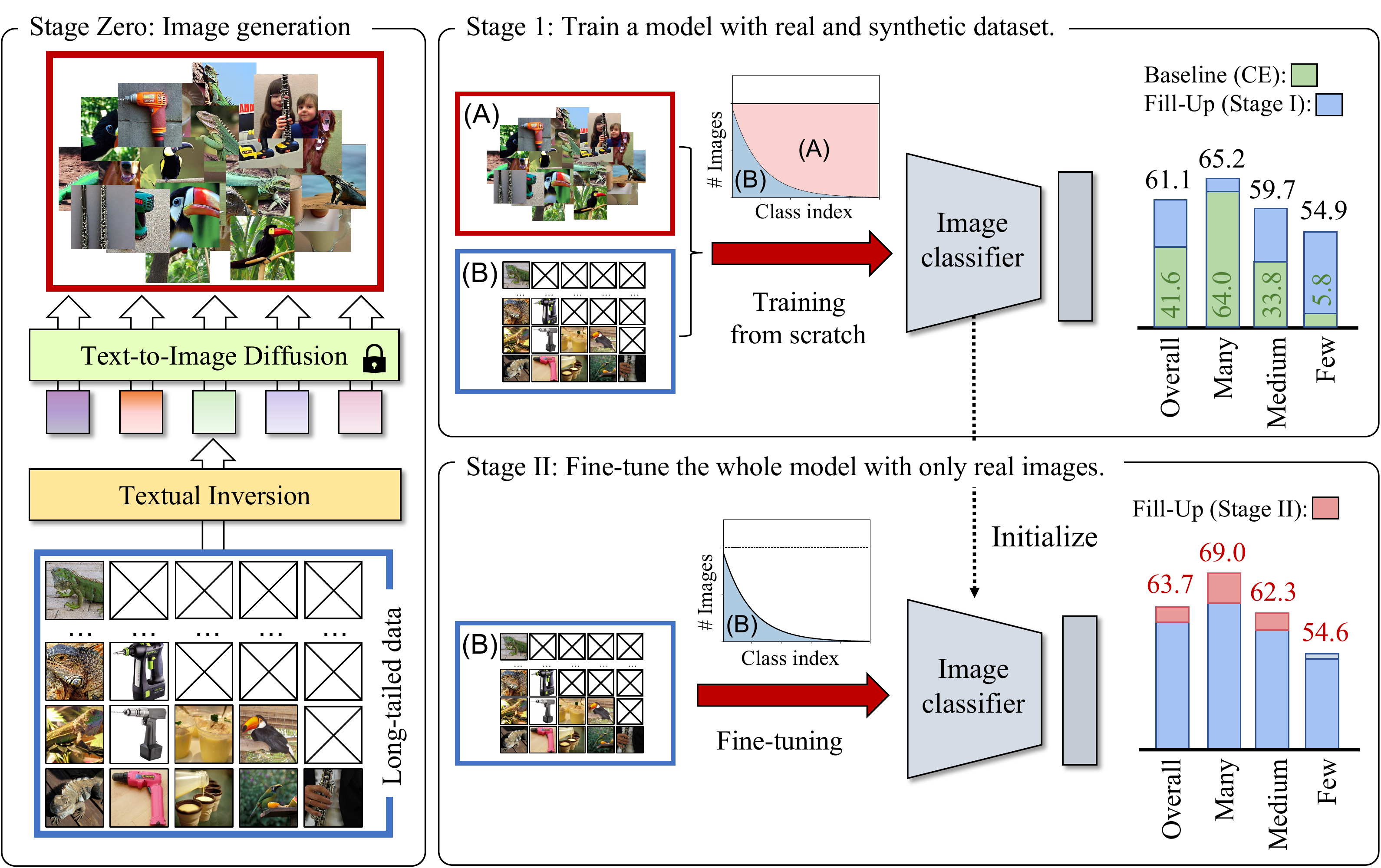}
    \caption{Overview of our proposed long-tailed image recognition pipeline on ImageNet-LT. During the Stage I, the image classifier is trained on balanced dataset, obtained by filling up long-tailed data with synthetic samples. Stage II fine-tunes the image classifier with long-tailed real dataset along with Balanced Softmax loss~\cite{ren2020balanced}.}
    \label{fig:framework}
    \vspace{-3mm}
\end{figure}
%

\subsection{Two-stage training procedure}
\label{sec:Method_2stage}
One of the mainstream approaches in long-tailed recognition is to decouple representation learning and classifier learning as suggested by Kang~\etal~\cite{Kang2020Decoupling}. Motivated by this idea, we adopt a two-stage training procedure. During the first stage, the entire model is trained on balanced dataset, obtained by filling up long-tailed data with synthetic samples. 

In the second stage, we fine-tune the model with only real samples on top of robust feature learned from the first stage~\cite{he2023is, sariyildiz2023fake}. Interestingly, while cRT~\cite{Kang2020Decoupling} only retrains the classifier~(\textit{i.e.} linear layer), fine-tuning the whole model with a reduced learning rate produced superior results in our case. For both stages, we adopt Balanced Softmax loss~\cite{ren2020balanced}, which effectively handles the discrepancy between the posterior distributions of training and test set. Given the number of samples in each class $i$ (denoted as $n_i$), and the model's output (denoted as $\eta$), Balanced Softmax~($\hat{\phi}$) can be expressed as  
    $\hat{\phi}_j =\frac{n_je^{\eta_j}}{\sum_{i =1}^k n_ie^{\eta_i}}$.
In both stages, only real samples are accounted for Balanced Softmax and no other re-sampling or re-weighting technique is adopted. The effectiveness of Balanced Softmax loss, which harmoniously blends with our method, is demonstrated through ablation experiments~in Sec.~\ref{sec:Experiments_Ablation}.
\vspace{-2mm}
\subsection{Comparison with existing methods}
\label{sec:Method_Comparison}
Here, we clarify the differences between our method and previous works~\cite{he2023is, sariyildiz2023fake, azizi2023synthetic} on synthetic data generation. He~\etal~\cite{he2023is} and Sariyildiz~\etal~\cite{sariyildiz2023fake} mainly focus on the zero-shot learning and transfer learning of the robust representation acquired from synthetic images. The key difference between these two is that Sariyildiz~\etal emphasize training the network from scratch, whereas He~\etal utilize synthetic data to fine-tune a pretrained CLIP model. Another notable work by Azizi~\etal~\cite{azizi2023synthetic} focus on tuning text-conditioned Imagen~\cite{saharia2022photorealistic} to a class conditional diffusion model and highlight their method using a metric called Classification Accuracy Score~\cite{ravuri2019classification}. 
We gently note that comparing the performance of this model, trained on full data, with the aforementioned zero-shot or few-shot methods may result in misleading conclusions. Our work primarily focuses on long-tailed scenarios through per-class optimization of random initialized text tokens. 

\section{Experiments}
\subsection{Datasets}
\label{sec:Experiments_Datasets}
\begin{wraptable}{r}{0.45\textwidth}
    \vspace{-0.45cm}
    \caption{Overview of long-tailed datasets}
    \label{dataset_overview}
    \resizebox{0.45\textwidth}{!}{
    \begin{tabular}{lccc}
    \cmidrule[1.0pt]{1-4}
    Dataset & \# Classes & \# Images & Imbalance Factor \\
    \cmidrule[1.0pt]{1-4}
    ImageNet-LT~\cite{liu2019large} & 1000  & 115.8K & 256 \\
    Places-LT~\cite{liu2019large} & 365 & 62.5K & 996 \\
    iNaturalist-LT~\cite{van2018inaturalist} & 8142 & 437K & 500 \\
    IN100-LT & 100 & 19K & 200 \\
    \cmidrule[1.0pt]{1-4}
    \end{tabular}}
    \vspace{-0.35cm}
\end{wraptable}
To validate our method, we conduct experiments on commonly used long-tailed datasets: ImageNet-LT~\cite{liu2019large}, Places-LT~\cite{liu2019large}, and iNaturalist2018~\cite{van2018inaturalist}. We utilize ImageNet-100 (IN100) and ImageNet-100-LT (IN100-LT) for our ablation studies. IN100 is a randomly selected subset of ImageNet-1K consisting of 115.8K training images, and IN100-LT is its long-tailed version with an exponentially decaying number of samples. Further details are summarized in Table~\ref{dataset_overview}. Our findings align with previous studies~\cite{saha2022backdoor, sariyildiz2023fake}, as we observe similar trends across ImageNet, ImageNet-LT, and their downscaled versions, IN100 and IN100-LT. It is important to note that direct comparison of IN100 results with previous papers~\cite{sariyildiz2023fake, bansal2023leaving} could lead to misleading results, as each study adopts a distinct set of randomly selected classes.

\begin{table}[ht!]
\vspace{-2mm}
\caption{Classification accuracy (\%) on ImageNet-LT and iNaturalist2018 using ResNet50 backbone. * denotes models trained with longer epochs (over 200). Our Fill-Up adopts longer training only for iNaturalist2018. We separately categorize ensembling-based methods involving multiple experts.}
\label{imagenetlt_inaturalist}
\centering
\vspace{1mm}
\resizebox{0.93\textwidth}{!}{
\begin{tabular}{lcccccccc}
\cmidrule[1.0pt]{1-9}
\multirow{2}*[-0.5ex]{\textbf{Method}}  &  \multicolumn{4}{c}{\textbf{ImageNet-LT}~\cite{liu2019large}} & \multicolumn{4}{c}{\textbf{iNaturalist2018}~\cite{van2018inaturalist}}\\
    \cmidrule[1.0pt]( r){2-5}
    \cmidrule[1.0pt]( l){6-9}
    & \text{Overall} & \text{Many} & \text{Medium} & \text{Few} & \text{Overall} & \text{Many} & \text{Medium} & \text{Few} \\
    \cmidrule[1.0pt]{1-9}
Baseline (CE)~\cite{park2021cmo}              & 41.6 & 64.0 &  33.8 & 5.8 & 61.0 & 73.9 & 63.5 & 55.5 \\ 
\midrule
Decouple-cRT~\cite{Kang2020Decoupling} & 47.3 & 58.8 & 44.0 & 26.1 & 68.2 &  73.2 & 68.8 & 66.1\\
Decouple-LWS~\cite{Kang2020Decoupling} & 47.7 & 57.1 & 45.2 & 29.3 & 69.5 &  71.0 &  69.8 &  68.8\\
Remix~\cite{chou2020remix}              & 48.6 & 60.4 & 46.9 & 30.7 & 70.5 & - & -& -\\
LDAM-DRW~\cite{cao2019learning}              & 49.8 & 60.4 & 46.9 & 30.7 & 66.1 & - & - & -\\
BS~\cite{ren2020balanced} & 51.0 & 60.9 & 48.8 & 32.1 & 70.0 & 70.0 & 70.2 & 69.9 \\
PaCo$^*$~\cite{cui2021parametric}            & 57.0 & 65.0 & 55.7 & 38.2  & 73.2 & 70.3 & 73.2 & 73.6\\ 
BS~+~CMO$^*$~\cite{park2021cmo} & 58.0 & 67.0 & 55.0 & 44.2 & 74.0 & 71.9 & 74.2 & 74.2 \\
\midrule
RIDE (4 experts)~\cite{wang2021longtailed} & 55.4 & 66.2 & 52.3 &  36.5 & 72.6 & 70.9 & 72.4 & 73.1\\
NCL (3 experts)$^*$~\cite{li2022nested}      & 59.5 & - & - & -  & 74.9 & 72.7 & 75.6 & 74.5\\
BalPOE (3 experts)$^*$~\cite{aimar2022balanced}  & 60.8 & 67.8 & 59.2 & 46.5 & \textbf{76.9} & \textbf{75.0} & \textbf{77.4} &  \textbf{76.9} \\

\midrule
Fill-Up (Stage I)         & 61.1 & 65.2 & 59.7 &\textbf{54.9} & 64.4 & 58.7 & 63.9 & 66.3 \\
Fill-Up (Stage II)$^{(*)}$    & \textbf{63.7} & \textbf{69.0} & \textbf{62.3} & 54.6 & 74.7 & 71.7 & 74.8 & 75.3 \\
\cmidrule[1.0pt]{1-9}
\end{tabular}}
\vspace{-3mm}
\end{table}
\begin{table}[ht]
\begin{minipage}[b]{0.542\linewidth}
    \caption{Classification accuracy (\%) on Places-LT trained using pretrained ResNet152. \textdagger~denotes models trained from scratch using ResNet50 backbone.}
    \label{places_lt}
    \vspace{1mm}
    \resizebox{0.95\textwidth}{!}{
    \begin{tabular}{lcccc}
    \cmidrule[1.0pt]{1-5}
      \multirow{2}*[-0.5ex]{\textbf{Method}}   & \multicolumn{4}{c}{\textbf{Places-LT}~\cite{liu2019large}}  \\
    \cmidrule[1.0pt]{2-5} & Overall & Many & Medium & Few  \\
    \cmidrule[1.0pt]{1-5}
    
    $\text{Baseline (CE)}^{\dagger}$    & 18.7 & 37.4 & 12.0 & 0.5 \\
    Baseline (CE)~\cite{cui2021parametric}    & 30.2 & 45.7 & 27.3 & 8.2 \\
    \midrule
    Decouple-LWS~\cite{Kang2020Decoupling} & 37.6 & 40.6 & 39.1 & 28.6 \\
    BS~\cite{ren2020balanced} & 38.6 & 42.0 & 39.3 & 30.5 \\
    ResLT~\cite{cui2022reslt} & 39.8 & 39.8 & 43.6 & 31.4 \\
    MiSLAS~\cite{zhong2021improving} & 40.4 & 39.6 & 43.3 & 36.1 \\
    PaCo~\cite{cui2021parametric} & 41.2 & 37.5 & \textbf{47.2} & 33.9 \\
    \midrule
    NCL (3 experts)~\cite{li2022nested} & 41.8  & - & - & - \\
    SHIKE (3 experts)~\cite{jin2023long} & 41.9 & 43.6 & 39.2 & \textbf{44.8} \\  
    \midrule
    $\text{Fill-Up (Stage I)}^{\dagger}$  & 40.0 & 41.5 & 41.3 & 34.5\\
    $\text{Fill-Up (Stage II)}^{\dagger}$ & 42.0 & 45.0 & 43.5 & 34.1  \\
    Fill-Up (Stage I) & 41.5 & 42.5 & 43.2 & 36.2 \\
    Fill-Up (Stage II) & \textbf{42.6} & \textbf{45.7} & 43.7 & 35.1 \\
    \cmidrule[1.0pt]{1-5}
    \end{tabular}}
\end{minipage}
\quad
\begin{minipage}[b]{0.41\linewidth}
    \caption{Classification accuracy (\%) on ImageNet-LT test set using different synthetic data generation methods.}
    \label{efficiency_table}
    \vspace{1mm}
    \resizebox{0.95\textwidth}{!}{
    \begin{tabular}{lcc}
    \cmidrule[1.0pt]{1-3}
    \multirow{2}*[-0.5ex]{\textbf{Method}}  & \multicolumn{2}{c}{\textbf{ImageNet-LT}~\cite{liu2019large}} \\
     \cmidrule[1.0pt]{2-3}
    & Guidance & Top-1\\
    \cmidrule[1.0pt]{1-3}
    ImageNet-LT & - & 41.6 \\
    \midrule
    Sariyildiz~\etal~\cite{sariyildiz2023fake} & 2.0 & 42.9 \\
    Bansal~\etal~\cite{bansal2023leaving} & 7.5 & 30.1 \\
    \midrule
    Single Template & 7.5 & 26.6 \\
    Single Template & 1.0 & 41.9 \\
    ~~~~~~+~ Real Data & - & 53.3  \\
    \midrule
    Textual Inversion (1.3K) & 1.0 & 49.9 \\
    ~~~~~~+~ Real Data & - & 55.6 \\
    ~~~~~~~~~~+~ BS + RandAug & - & 61.8 \\
    Textual Inversion (2.6K) & 1.0 & 53.5 \\
    ~~~~~~+~ Real Data & - & 57.1 \\
    ~~~~~~~~~~+~ BS + RandAug & - & 63.7 \\
    \midrule
    ImageNet (Full) & - & 76.1 \\
    \cmidrule[1.0pt]{1-3}
    \end{tabular}}
\end{minipage}
\vspace{-3.5mm}
\end{table}

\subsection{Implementation details}
\label{sec:Experiments_Implementation}
For generating synthetic samples, we adhere to our best setting from Sec.~\ref{sec:Method_ti}. To facilitate comparison, we utilize the widely adopted ResNet50~\cite{He_2016_CVPR} for all experiments. For Places-LT, we additionally report scores initiated from pretrained ResNet152~\cite{He_2016_CVPR}, following the conventional approaches~\cite{yang2022survey, zhang2023deep}. With RandAugment~\cite{cubuk2020randaugment}, our models are trained for 100 epochs in Stage I and fine-tuned for 30 epochs in the Stage II. However, we make an exception for the fine-grained iNaturalist2018 dataset, extending the Stage II to 400 epochs, based on the notable performance increase observed with longer training schedules, as recommended by~\cite{cui2021parametric}. After evaluating the models on the corresponding balanced test/validation sets, we report overall top-1 accuracy as well as accuracies for many-shot (more than 100 images), medium-shot (20-100 images), and few-shot (less than 20 images) classes. Further details are available in Appendix.

\vspace{-2mm}
\subsection{Efficiency of Textual Inversion generated images}
\label{sec:Experiments_Efficiency}
Before delving into long-tailed benchmarks, we first evaluate the efficacy of the Textual Inversion-based generation strategy on the scale of ImageNet. Table~\ref{efficiency_table} presents the results of Stage I training with synthetic data targeted for ImageNet-LT. Following standard ImageNet data, which contains 1,300 images for each class, we generate either 1,300 or 2,600 images, resulting in a total of 1.3M and 2.6M images, respectively. If real data is provided, we balance it by filling it up to contain 1,300 or 2,600 images per class accordingly. Aligning with our observations on IN100 and IN100-LT, generating images from textual-inverted tokens yields 8\% higher top-1 accuracy than using single template generated data when trained solely with the synthetic data, and 2.3\% higher top-1 accuracy when trained with filled-up real data. We also observe that these numbers scale as the amount of synthetic samples grow. Although it is important to mention that our method uses long-tailed real samples to learn text-tokens, it is noteworthy that this approach outperforms the previous state-of-the-art generation method from Sariyildiz~\etal~\cite{sariyildiz2023fake} which employs DINO style data augmentation, by 7\% without employing any data augmentation, using the same number (1,300) of images. 

\vspace{-3mm}
\subsection{Benchmarks on long-tailed datasets}
\label{sec:Experiments_Benchmarks}
We compare the proposed two-stage method based on Textual Inversion with the state-of-the-art methods. The benchmark results for ImageNet-LT and iNaturalist2018 can be found in Table~\ref{imagenetlt_inaturalist}, and results for Places-LT are displayed in Table~\ref{places_lt}. The benchmark results from ImageNet-LT demonstrate that our method effectively learns a balanced representation in Stage I, and achieves a notable performance improvement by learning decisive features from the real domain in Stage II. Especially, our method exhibits a significantly higher top-1 accuracy for few-shot classes, demonstrating more than 16\% performance gain compared to PaCo~\cite{cui2021parametric}, which previously achieved the highest accuracy for few-shot classes. 

iNaturalist2018 dataset is not only known for its imbalanced data distribution, but also for its extremely fine-grained classes~\cite{yang2022survey}. The relatively modest performance gain observed during our Stage I training can be attributed to the noise in the data generated using a low classifier-free guidance of scale 1.0. However, despite the potential hindrance caused by noisy samples in learning fine-grained classes, our Stage II model, trained on top of balanced feature space derived from Stage I, exhibits a substantial performance improvement while retaining balanced knowledge. We also note remarkable performance gain from 70.4\% to 74.7\% with PaCo-style longer training schedules.

Places-LT benchmark, as displayed in Table~\ref{places_lt}, is a very challenging dataset where training of ResNet50 from scratch only yields 18.7\% overall accuracy and an exceptionally low 0.5\% accuracy for few-shot classes. As a result, adopting an ImageNet pretrained ResNet152 is a common practice. However, our proposed method not only outperforms existing state-of-the-art approaches with the standard protocol of ResNet152 but also surpasses them when trained from scratch using ResNet50. 

\begin{table}[ht]
\begin{minipage}[t]{0.475\linewidth}
    \centering
    \captionof{figure}{An illustration of different Fill-Up methods. Long-tailed data represents the actual class distribution of IN100-LT.}
    \includegraphics[width=0.98\linewidth]{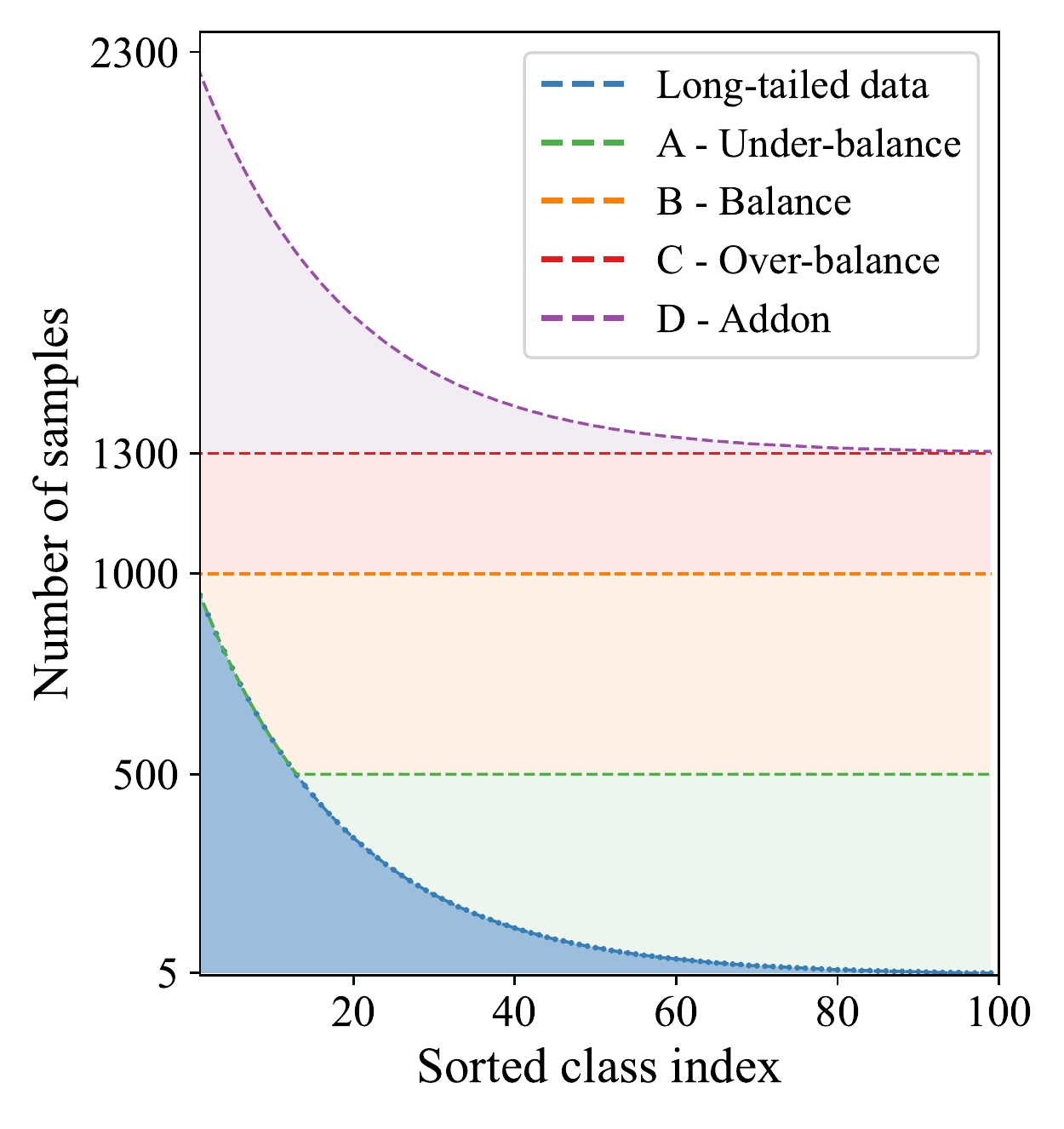}
    \label{ablation_figure}
\end{minipage}
\quad
  \begin{minipage}[t]{0.475\textwidth}
    \centering
    \caption{Classification accuracy (\%) on IN100-LT using ResNet50. Diverse Fill-Up approaches are denoted in accordance with Figure~\ref{ablation_figure}.}
    \vspace{1mm}
    \resizebox{1.0\textwidth}{!}{
    \begin{tabular}{lcccc}
    \cmidrule[1.0pt]{1-5}
      \multirow{2}*[-0.5ex]{\textbf{Method}}   & \multicolumn{4}{c}{\textbf{IN100-LT}}  \\
    \cmidrule[1.0pt]{2-5} & Overall & Many & Medium & Few  \\
    \cmidrule[1.0pt]{1-5}
    Baseline (Full Data)    & 82.3 & 81.3 & 84.1 & 81.8 \\
    Baseline (LT data)    & 26.4 & 50.8 & 13.5 & 1.9 \\
    ~~+~BS  & 29.3 & 45.1 & 27.5 & 6.1 \\
    ~~+~BS + RandAug & 27.9 & 43.3 & 26.7 & 4.7\\
    Fake Only &66.1 & 67.0 & 71.5 & 56.7 \\
    \midrule
    Stage I    & &  &  &  \\
    \midrule
    ~~+~A (Under-balance) & 56.0 & 70.7 & 53.6 & 35.0\\
    ~~+~B (Balance) & 65.8 & 77.3 & 65.9 & 47.2 \\
    ~~+~C (Over-balance) & 69.5 & 80.1 & 70.2 & 52.0\\
    ~~~~~+~BS & 71.2 & 73.0 & 73.1 & 66.1 \\
    ~~~~~+~BS + RandAug & 72.0 & 73.8 & 72.9 & 68.2 \\
    ~~+~D (Addon) & 70.1 & 81.1 & 70.9 & 51.7\\
    \midrule
    \multicolumn{5}{l}{Stage II}  \\
    \midrule
    Stage I (C+~BS+~RandAug) & 69.5 & 80.1 & 70.2 & 52.0 \\
    ~~+~Na\"ive &67.8 & 78.8 & 66.4 & 51.7\\
    ~~+~Class-balanced sampling & 64.4 & 75.3 & 67.0 & 44.1 \\
    ~~+~cRT & 72.7 & 79.7 & 76.9 & 56.8\\
    ~~+~BS & 74.0 & 79.1 & 75.9 & 63.8\\
    ~~~~~+~RandAug & 74.5 & 79.8 & 77.0 & 63.1\\
    \cmidrule[1.0pt]{1-5}
    \end{tabular}}
    \label{ablation_table}
    \end{minipage}
    \vspace{-5mm}
\end{table}

\subsection{Ablation studies}
\label{sec:Experiments_Ablation}
\vspace{-1.2mm}
Here, we perform extensive ablations using IN100 and IN100-LT datasets, focusing on the impact of different Fill-Up strategies and the effectiveness of our design choice in two-stage training. As depicted in Fig~\ref{ablation_figure}, we adopt four distinct strategies to fill the long-tailed data with synthetic images. In line with the previous observation from Fig~\ref{fig:imgb}, performance generally tends to scale with an increasing number of images used for Fill-Up. However, experimental results suggest that na\"ively adding all 1,300 images for every class (referred to as "D - addon") without considering the relative balance among classes leads to only marginal performance gains. It is also important to highlight the powerful synergy between our method and the Balanced Softmax loss. Compared to the baseline, the accuracy of the few-shot classes rises from 6.1\% to an impressive accuracy of 66.1\%.

For Stage II training, we begin from our best setting in Stage I. As suggested by previous works, na\"ive fine-tuning or class-balanced sampling degrades the balanced features learned from the first stage under extremely long-tailed data. cRT-style fine-tuning of the final classifier brings 3.2\% performance gains on top of the balanced features. In our case, fine-tuning the whole model with Balanced Softmax loss resulted in a stronger performance. We believe, this unbiased extension of the softmax function, effectively mitigates the challenges posed by highly imbalanced data, allowing the entire set of model parameters to be efficiently tuned without collapsing. Given the additional 0.5\% gain from RandAugment, our method achieves a remarkable accuracy of 74.5\% which significantly reduces the gap with real data (82.3\%) compared to the long-tailed baseline (26.4\%).
\vspace{-1mm}

\section{Conclusion}
\vspace{-2.5mm}
This paper introduces the concept of filling-up long-tailed data with synthetic samples from recent large-scale text-to-image synthesis models. We provide valuable insights into different methods of synthetic data generation through detailed experiments and propose a novel approach leveraging Textual Inversion that outperforms other methods. Carefully designed with real-world data imbalance scenarios in mind, our approach, which comprises per-class optimization of text tokens and two-stage training, displays state-of-the-art results on standard long-tailed benchmarks, with particularly notable performance in few-shot classes accuracy. 
\vspace{-2.0mm}
\paragraph{Limitations and future work.} While synthetic data generation demonstrates highly scaled performance with an increasing number of generated images, vast number of sampling through diffusion process demands massive computational resources. Recent advancements in one-step generative models (\textit{e.g.} GAN~\cite{kang2023gigagan, sauer2023stylegan}, and consistency models~\cite{song2023consistency}) offer potential solutions. Furthermore, the ability to generate unlimited data comes at the cost of increased computation for training. While our approach shows higher efficiency in generating feature-rich samples, the challenge of generating images with strong features on par with real samples remains as another core problem for future.

\newpage
{\small
    \bibliographystyle{unsrtnat}
    \bibliography{main}
}

\clearpage
\appendix
\addcontentsline{toc}{section}{Appendices}
\renewcommand{\thefigure}{A\arabic{figure}}
\renewcommand{\thetable}{A\arabic{table}}
\section*{\Large{Appendices}}
\section{Diffusion models}
\label{sup_diffusion}
\subsection{Denoising diffusion probabilistic model}
In this section, we provide a basic explanation of Denoising Diffusion Probabilistic Model (DDPM)~\cite{Ho2020DenoisingDP} and Latent Diffusion Model~(LDM)~\cite{rombach2022high}. Diffusion model involves thermodynamics-inspired processes, namely the forward and reverse processes, which are associated with a Markov chain with Gaussian transition to gradually add and remove noises. Specifically, each step of the forward process (also called the diffusion process) is defined by a conditional Gaussian distribution where $\beta_{t}$ controls mean and variance of output noises. It is demonstrated that $q(x_{T})$ follows an isotropic standard Gaussian distribution under reasonable $T$ and $\beta_{t}$ constants. The reverse process, on the other hand, aims at recovering the original data distribution $q(x_{0})$ from the isotropic standard Gaussian distribution $\mathcal{N}(0, \text{I})$. Each reverse process is defined by a conditional Gaussian distribution $p_{\theta}(x_{t-1} \vert x_t)$, where the mean and variance are determined by a neural network parameterized by $\theta$ and time-dependent constants $\sigma_{t}$, respectively. Mathematically, each step of the forward and the reverse processes can be expressed as follows:
\begin{align*}
    q(x_{t}|x_{t-1}) &:= \mathcal{N}(x_{t}; \sqrt{1 - \beta_{t}}x_{t-1}, \beta_{t}\text{I}), \label{eq:seq1}\tag{A1}\\
    p_{\theta}(x_{t-1}|x_{t}) &:= \mathcal{N}(x_{t-1}; \mu_{\theta}(x_{t}, t), \sigma_{t}^{2}). \label{eq:seq2}\tag{A2}
\end{align*}
Since the reverse process (Eq.~(\ref{eq:seq2})) is not analytically derived from the forward process~(Eq.~(\ref{eq:seq1})), current diffusion models approximate the true conditional distribution for reverse step $q(x_{t-1}|x_{t})$ by plugging in $p_{\theta}(x_{t-1}|x_{t})$ instead of $q(x_{t-1}|x_{t})$ and minimizing negative log-likelihood for the true data distribution. Since calculating the exact log-likelihood is intractable, Ho~\etal~\cite{Ho2020DenoisingDP} optimize a variational lower bound $\mathcal{L}_{\text{VLB}}$ on the negative log-likelihood. The variational lower bound is expressed in Eq.~(\ref{eq:eq3}).
\begin{align*}
    \mathcal{L}_{\text{VLB}} & := \mathbb{E}_{q}\bigg[-\log\frac{p_{\theta}(x_{0:T})}{q(x_{1:T}|x_{0})}\bigg], \label{eq:eq3}\tag{A3} \\
    & = \mathbb{E}_{q}\bigg[\underbrace{\text{KL}(q(x_T|x_{0})~\Vert~p(x_{T})\big)}_{\mathcal{L}_{T}} + \sum_{t > 1}\underbrace{\text{KL}\big(q(x_{t-1}|x_{t}, x_{0})~\Vert~p_{\theta}(x_{t-1}|x_{t}))}_{\mathcal{L}_{t-1}} \underbrace{- \log p_{\theta}(x_{0}|x_{1})}_{\mathcal{L}_{0}}\bigg].
\end{align*}
The above $\mathcal{L}_{\text{VLB}}$ consists of three sub losses: 1) loss for the forward process $\mathcal{L}_{T}$, 2) loss for the backward process $\mathcal{L}_{t-1}$, and 3) discrete decoder loss $\mathcal{L}_{0}$. Although the above variational lower bound can be computed analytically (KL divergence between two Gaussian distributions has an exact solution), the authors of DDPM utilize the practical loss function below and demonstrate the successful generation of realistic images from the real-world domain.
\begin{align*}
    \mathcal{L}_{\text{simple}}(\theta) := \mathbb{E}_{x_{0} \sim q(x), \eps \sim \mathcal{N}(0, \text{I}), t}\Big[\big\Vert\eps - \eps_{\theta}\big(\sqrt{\bar{\alpha_{t}}} x_{0} + \sqrt{1 - \bar{\alpha_{t}}}\eps, t\big) \big\Vert^{2}_{2}\Big],
    \label{eq:eq4}\tag{A4}
\end{align*}
where $\bar{\alpha_{t}} = \prod_{s=1}^{t}\alpha_{t}.$ 

Finally, image sampling is conducted by recursively iterating the following equation:
\begin{align*}
    x_{t-1} = \frac{1}{\sqrt{\alpha_{t}}}\Big(x_{t} - \frac{1-\alpha_{t}}{\sqrt{1-\bar{\alpha_{t}}}} \eps_{\theta}(x_{t}, t)\Big) + \sigma_{t}z, \quad\quad \text{where} \; z \sim \mathcal{N}(0, \text{1}). \label{eq:eq5}\tag{A5}
\end{align*}

\subsection{Latent diffusion model}
\label{sup_ldm}
Latent Diffusion Model~(LDM) reduces the excessive training and inference time of DDPM by performing forward and reverse processes in the latent space. Specifically, an encoder $\mathcal{E}$ learns to map an arbitrary image $x \sim q(x)$ into a latent code $z=\mathcal{E}(x)$, and a decoder $\mathcal{D}$ learns to map such latent back into the image space ($\mathcal{D}(\mathcal{E}(x))\approx x$). Regularization by either KL-divergence loss~\cite{van2017neural} or VQGAN-like~\cite{esser2021taming} vector quantization is adopted to avoid high-variance latent spaces. Given time step $t$, diffused latent variable $z_t$, unscaled noise sample $\epsilon$, denoising network $\epsilon_\theta$, conditioning network $c_{\theta}$, and conditional input $y$, the LDM loss is given as follows:
\begin{align*}
    \mathcal{L}_{\text{LDM}} := \mathbb{E}_{z \sim \mathcal{E}(x), y, \epsilon \sim \mathcal{N}(0, 1), t}{\Big[ \Vert \epsilon - \epsilon_{\theta}( z_{t},t, c_{\theta}(y)) \Vert_{2}^{2}\Big]}
    \label{eq:eq6}\tag{A6}
\end{align*}
Stable Diffusion is a specific text-conditioned variant of the LDM trained on LAION-5B~\cite{schuhmann2022laion} dataset which utilizes CLIP text encoder~\cite{radford2021learning} as a conditional network $c_{\theta}$ for text input $y$. In LDM variants, classifier-free guidance~\cite{ho2021classifierfree} simultaneously guides the reverse process by calculating the difference between the text- and null-conditioned score functions. The mathematical formulation of classifier-free guidance can be written as follows:
\begin{align*}
    \tilde{\epsilon}_{t} &= \epsilon_{\theta}(z_{t}) + w(\epsilon_{\theta}(z_{t}, c) - \epsilon_{\theta}(z_{t})), 
    \label{eq:eq7}\tag{A7}
\end{align*}
where $\tilde{\epsilon}_{t}$ is a guided output, and $w$ is a guidance scale.

\section{Additional experimental details}
\label{sup_experiment}

\subsection{Details on prompt-to-image based methods}
\label{sup_p2i}
\begin{wraptable}{r}{0.5\textwidth}
\vspace{-0.45cm}
\caption{Effect of different decoding strategies on Top-1 and Top-5 accuracy.}
\label{sup_decoding}
\centering
\resizebox{0.5\textwidth}{!}{
    \begin{tabular}{lccc}
    \cmidrule[1.0pt]{1-4}
    \textbf{Decoding Method} & \text{T5} & \text{Flan-T5 XXL} & \text{Prompt-Extend} \\ 
    \cmidrule[1.0pt]{1-4}
    Beam Search & 34.2 / 61.7 & 37.0 / 65.7 & 27.1 / 50.9 \\ 
    Top-p Sampling & 36.0 /	61.0 & 41.3 / 67.3 & 35.0 / 58.8 \\
    \cmidrule[1.0pt]{1-4}
    \end{tabular}
}
\end{wraptable}

For prompt-to-image-based generation, we employ five different methods: \textit{single template, CLIP templates, T5, Flan-T5 XXL, and Prompt-Extend}. In the previous work by He~\etal~\cite{he2023is}, they apply CLIP filtering to refine the outputs generated by the T5 language model, aiming to remove noisy prompts. However, we believe it is worth considering the noisy prompts as they are an inherent part of the language model and have potential value in evaluating the model's performance. Hence, we refrain from applying additional filters to the generated prompts. 

Among a wide range of T5~\cite{raffel2020exploring} variants fine-tuned on different datasets, we specifically utilize T5 model fine-tuned on CommonGen dataset~\cite{lin-etal-2020-commongen, t5-huggingface}. Consistent with the findings from He~\etal~\cite{he2023is}, we observe prompts generated from this variant tend to generate natural descriptions of plausible real-world objects. We supply the model with the class name and let it generate a sentence up to 60 tokens long. 

For a larger model, we adopt Flan-T5 XXL~\cite{chung2022scaling}, which is an instruction-tuned variant of T5 XXL. Specifically, we utilize the model fine-tuned through the recently popular Stanford Alpaca~\cite{alpaca, flan-t5-huggingface}. Our specific instruction for the model is as follows: "Generate a prompt-style sentence for image generation including \{\texttt{CLS}\}". In general, this approach led to the generation of significantly more plausible outputs with reduced noise. 

We observe that not all generated prompts can be effectively described by the Stable Diffusion due to limitations in its expressiveness. Motivated by this observation, we employ a widely used prompt-extender~\cite{prompt-extend-huggingface} for Stable Diffusion, which is a GPT-2 model fine-tuned on the DiffusionDB~\cite{wang2022diffusiondb} dataset. Like T5, we supply the model with the class name and let the model extend prompts in a more favorable way for Stable Diffusion.

For language models, improved decoding methods also play a crucial role in enhancing the quality of the outputs. To evaluate their performances, we conduct experiments using two widely used decoding techniques: beam search and top-p nucleus sampling. To obtain $n$ prompts, we use $4\times n$ beams for beam search and use $k = 50$ and $p = 0.95$ for sampling. We generate 200 and 1,000 prompts for beam search and top-p nucleus sampling, respectively. Prompts are evenly utilized to generate 1,300 images per class. Following previous experiments, we generate 130K images for IN100 with a scale of 7.5. As seen in Table~\ref{sup_decoding}, we generally observe better performance with top-p sampling. The final table for prompt-to-image-based methods, covering scales of 1.0 and 7.5, is given in Table~\ref{sup_p2i_table}.

\begin{table}[ht!]
\vspace{-2mm}
\caption{Full comparison of accuracies among different prompt-to-image-based generation methods for IN100. Low Guidance indicates a scale of 1.0, and high guidance denotes 7.5.}
\label{sup_p2i_table}
\centering
\vspace{1mm}
\resizebox{0.9\textwidth}{!}{
\begin{tabular}{lccccc}
\cmidrule[1.0pt]{1-6}
& \multicolumn{5}{c}{\textbf{Prompt-to-Image}} \\
\cmidrule[1.0pt]( r){2-6}
\textbf{Method (Top-1/Top-5)} & \text{Single Template} & \text{CLIP Templates} & \text{T5} & \text{Flan-T5 XXL} & \text{Prompt-Extend} \\
\cmidrule[1.0pt]{1-6}
Low Guidance & 59.9 / 83.7   &  57.2 / 82.2      & 45.8 / 71.5    & 51.6 / 77.3 &  29.7 / 54.7 \\
High Guidance & 41.5 / 67.5   &  48.7 / 73.6      & 36.0 / 61.0    & 41.3 / 67.3 & 35.0 / 58.8 \\

\cmidrule[1.0pt]{1-6}
\end{tabular}}
\vspace{-2mm}
\end{table}

For higher guidance, which better reflects the complexity of given prompts, incorporating diversity in prompts seems to improve accuracy. However, additional complexity from fancy prompts tends to deteriorate the recognition ability in lower guidance situations with more noise engaged. This motivates our intuition of using Textual Inversion and adopting a simple prompt ("a photo of *") with a more context-rich token ("*"). Furthermore, using specific domain-favored prompts, such as Prompt-Extend favored by Stable Diffusion, can lead to greater deviation from the original distribution. Qualitative results for all generated prompts and images can be found in Sec.~\ref{sup_qualitative_generated_prompts} and Sec.~\ref{sup_qualitative_generated_images}.

\subsection{Details on image-to-image based methods}
\label{sup_i2i}
We directly utilize models from~\cite{sd1-variation, sd2-reimagine} for image-to-image-based methods. Since these two have different training setups, they have distinct default guidance scales, which are 3.0 for \textit{Image Variation} and 10.0 for \textit{Reimagine}. Reimagine also has a parameter called noise level, which controls the amount of noise added to the image embedding. Nonetheless, the impact of this parameter is minimal in top-1 accuracy (41.6\% with noise level 0 and 41.3\% with noise level 100 in a scale of 10.0), so we keep the noise level to 0. Moreover, considering that the Stable Diffusion 2 Reimagine model produces images with a resolution of 768 pixels, we note a potential advantage of this variant over all other approaches that generate images with a resolution of 512 pixels. When generating synthetic images, we try to maintain a balanced utilization of the given source images. For example, to generate 1,300 images using 5 real samples in the tail class, we utilize each source image 260 times for image-to-image generation. As discussed in the main paper, image-to-image-based methods sometimes tend to replace a crucial class-related component of an image with a "reimagined" object. Qualitative results of image-to-image based methods, including mentioned circumstances, can be found in Sec.~\ref{sup_qualitative_generated_images}.

\subsection{Details on transmodal methods}
\label{sup_transmodal}
For transmodal methods, we introduce two novel approaches: \textit{Captioning and Textual Inversion}. For captioning model, we utilize 6.7B parameter variant of BLIP2~\cite{li2023blip}. We extract a single prompt using a default greedy decoding method for every single image in the long-tailed data. Unlike previous prompt-to-image-based approaches, we observe prompts diverging with more advanced decoding methods, failing to describe the images accurately. We evenly distribute prompts and use them to generate 1,300 images per class. 

In the case of the \textit{Textual Inversion} method, we optimize a single token "*" of embedding shape 4$\times$768 for every class unless otherwise specified. To avoid manual handcrafted optimization, we adopt a unified learning rate of $5.0\times10^{-3}$, batch size of 2, and a simple heuristic for determining the total training steps, given by~($\min(\max(\textit{number of images}\times100, 2000), 10000)$). We save checkpoints every 100 steps and evenly utilize all of them for generating images. From experimental results, we believe that incorporating coarse-to-fine details during the optimization process contributes to learning robust class representations. In terms of storage, even for the head classes, the file sizes strictly remain below 1.6MB per class due to our limitation on the maximum training steps.

\subsection{Details on training classifiers}
\label{sup_classifier}
For all Stage I training, we follow the standard ResNet50 training protocol for ImageNet. We randomly crop the images to a size of 224$\times$224 and perform random horizontal flipping and color jittering. We train for 100 epochs with a batch size of 256 and an initial learning rate of 0.1, which is decayed by a factor of 0.1 every 30 epochs. If noted with RandAug or BS, standard RandAugment~\cite{cubuk2020randaugment} or Balanced Softmax loss~\cite{ren2020balanced} was adopted following Park~\etal~\cite{park2021cmo}. For Stage II training, we continue training for 30 epochs from the last checkpoint in Stage I with only real data. During the second stage, the initial learning rate of 1$\times10^{-3}$ decayed by a factor of 0.1 every ten epochs is adopted with five warm-up epochs. However, as noted in the main paper, we adopt the PaCo~\cite{cui2021parametric} style (400 epochs and initial learning rate of 0.01 decayed by cosine schedule) Stage II training for iNaturalist2018 as it outperforms our basic 30 epoch training by a huge margin. Lastly, to accelerate training for iNaturalist2018, which becomes a huge dataset containing over 8M images with our balanced Fill-Up strategy, we specifically utilize a batch size of 2,048 for iNaturalist2018.

\section{Extended analysis}
\label{sup_analysis}
\subsection{Alignment of generated data with real data}
Here, we discuss the alignment between generated data and real data. One of the most famous metrics to quantitatively measure the quality of generated images in relation to the true distribution is the Fr\'echet Inception Distance(FID)~\cite{Heusel2017GANsTB}. FID calculates the distance between two distributions in the feature space provided by a pre-trained Inception-V3~\cite{Szegedy2016RethinkingTI} network. We also utilize Precision \& Recall~\cite{Kynknniemi2019ImprovedPA}, which are widely adopted in the field of generative modeling, to quantify the fidelity and the diversity of the generated images separately. Precision is defined as the portion of generated images that falls within the support of the real data distribution, while recall is defined as the portion of real images falling within the support of the fake distribution~\cite{kang2022studiogan}. We adopt Inception-V3 network as a feature extractor for Precision \& Recall. In line with previous experiments, we generate a total of 130K images for IN100, with 1,300 images per class. Moreover, we maintain the practice of utilizing IN100-LT exclusively for learning text tokens and do not mix real samples during the evaluation.

\begin{figure}[ht]
    \centering
    \includegraphics[width=1.0\linewidth]{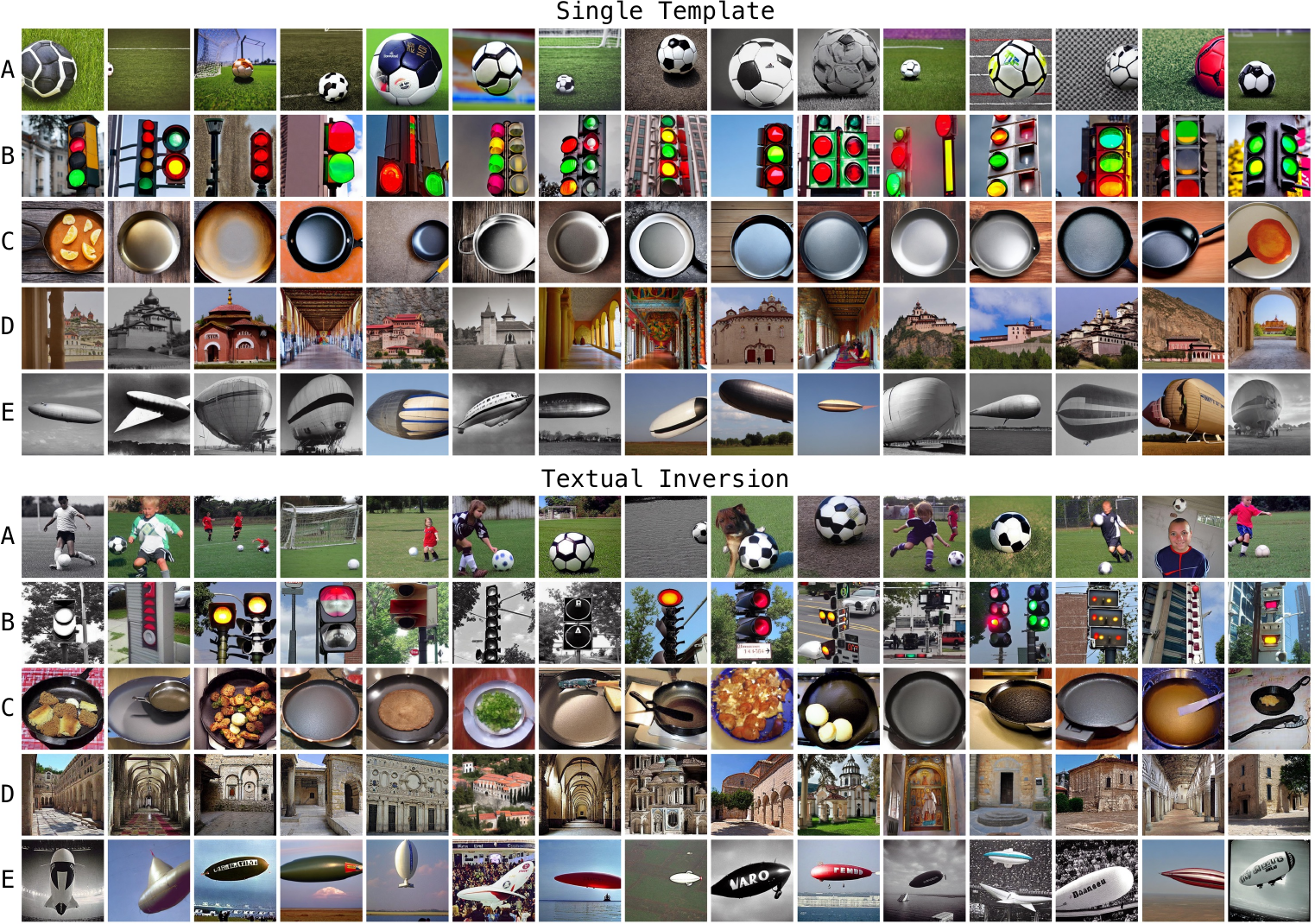}
    \caption{Randomly selected images from IN100 classes generated via Single Template and Textual Inversion methods. For better visualization, images were generated using a guidance scale of 7.5. Class names are as follows: \{A: soccer ball, B: traffic light, C: frying pan, D: monastery, E: airship\} In general, Textual Inversion produces more diverse images.}
    \label{singlevsti}
\end{figure}

As indicated in Table~\ref{fid_pr}, we generally observe better FID and Precision \& Recall values with Textual Inversion-generated images. A qualitative comparison between these two methods is depicted in Fig.~\ref{singlevsti}. In general, textual-inverted tokens exhibit key features from the real domain, enabling the generation of images with some diversity similar to real samples. Moreover, we confirm that classifier-free guidance plays a crucial role in maintaining a balance between Precision \& Recall. Although it is still very difficult to directly correlate these metrics with Top-1 accuracy, we suspect that achieving a high Recall value is important for improving accuracy. More qualitative analysis from Sec.~\ref{sup_qualitative_generated_images} supports the notion that noisier samples, generated with a low guidance scale, sparsely contain feature-rich instances that aid in the gradual expansion of the decision boundary. 

\begin{table}[h]

\caption{Comparison of generated images via single template method and Textual Inversion method on IN100. Real samples from IN100-LT are used to optimize text tokens. FID and Precision \& Recall are calculated against 130K real images from IN100.}
\centering
\vspace{1mm}
\resizebox{0.88\textwidth}{!}{
\begin{tabular}{ccccccccc}
\cmidrule[1.0pt]{1-9}
 &  \multicolumn{4}{c}{\textbf{Single Template}} & \multicolumn{4}{c}{\textbf{Textual Inversion}}\\
    \cmidrule[1.0pt]( r){2-5}
    \cmidrule[1.0pt]( l){6-9}
Scale & \text{Top-1~(\%)}~$\uparrow$ & \text{FID}~$\downarrow$ & \text{Precision}~$\uparrow$ & \text{Recall}~$\uparrow$ & \text{Top-1~(\%)}~$\uparrow$ & \text{FID}~$\downarrow$ & \text{Precision}~$\uparrow$ & \text{Recall}~$\uparrow$ \\
\cmidrule[1.0pt]{1-9}
0.0 & 1.3 & 130.2 & 0.292 & 0.632 & 1.3 & 130.2 & 0.292 & 0.632 \\
0.5 & 37.0 & 130.3 & 0.206 & 0.751 & 40.5 & 133.3 & 0.210 & 0.751 \\
1.0 & 59.9 & 75.2 & 0.329 & 0.670 & 66.1 & 77.5 & 0.338 & 0.691 \\
1.1 & 57.8 & 65.9 & 0.362 & 0.658 & 66.4 & 68.2 & 0.378 & 0.670 \\
1.5 & 61.4 & 42.4 & 0.492 & 0.577 & 65.9 & 41.7 & 0.503 & 0.607 \\
2.0 & 58.0 & 29.3 & 0.598 & 0.498 & 66.6 & 26.0 & 0.600 & 0.536 \\
3.0 & 55.1 & 21.6 & 0.703 & 0.386 & 63.1 & 15.0 & 0.695 & 0.450 \\
5.0 & 45.4 & 21.1 & 0.764 & 0.254 & 54.9 & 12.3 & 0.743 & 0.350 \\
7.5 & 41.5 & 22.8 & 0.771 & 0.180 & 47.2 & 13.4 & 0.744 & 0.290 \\
9.0 & 37.5 & 23.9 & 0.765 & 0.160 & 47.6 & 14.0 & 0.737 & 0.266 \\
\cmidrule[1.0pt]{1-9}
\label{fid_pr}
\vspace{-8mm}
\end{tabular}}
\end{table}

\subsection{Per class accuracies of Fill-Up}
\begin{figure}[t!]
    \centering
    \includegraphics[width=0.95\linewidth]{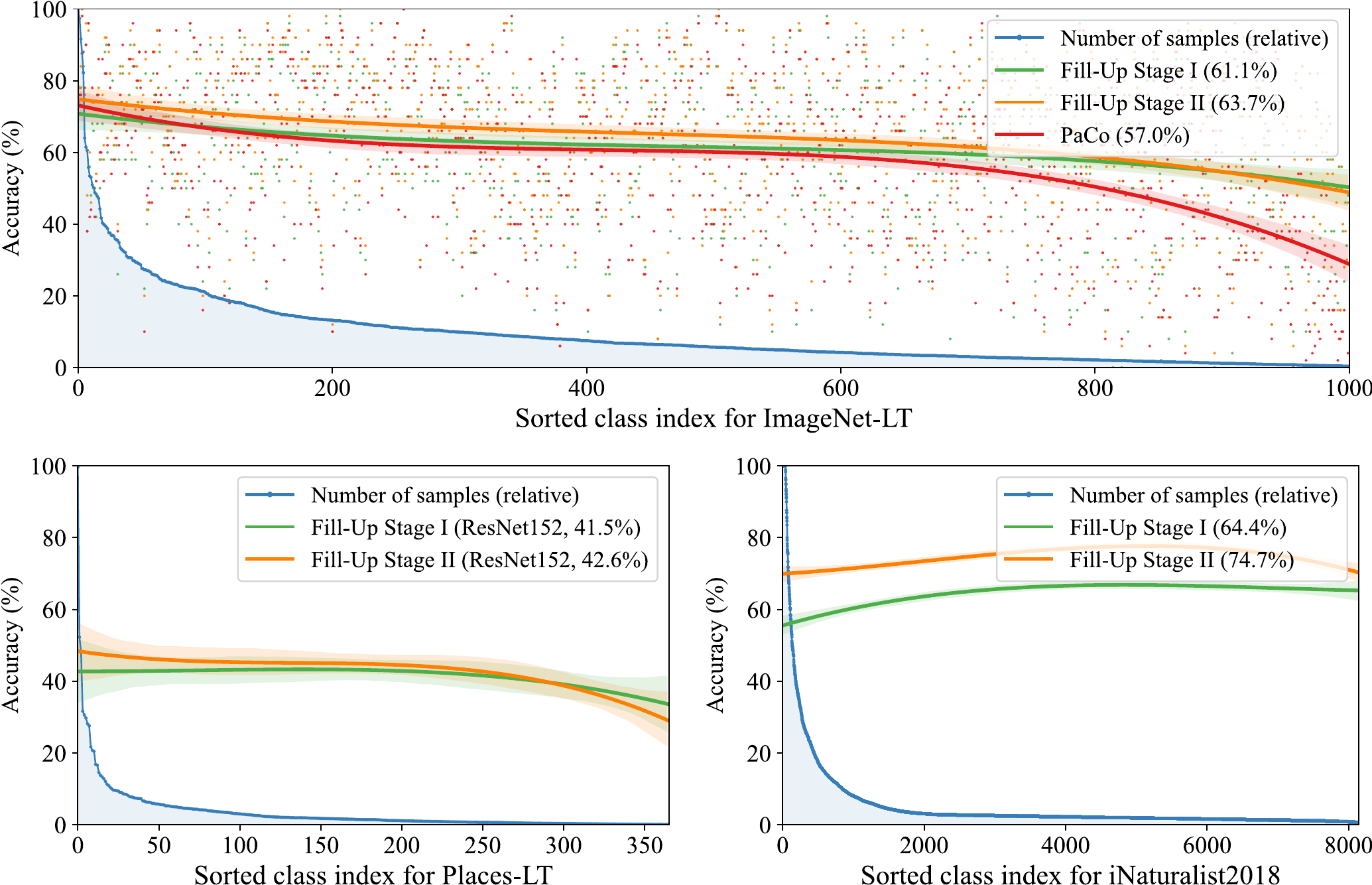}
    \caption{Comparison of per-class accuracies using different methods. Colored lines indicate polynomial regression graphs of order 3, with bands indicating the 95\% confidence interval. For ImageNet-LT, colored dots represent the actual values. We do not include dots for Places-LT and iNaturalist2018 as their evaluation set consists of only 20 and 3 images per class, respectively, resulting in a flat line of no significant variation. Fill-Up generally presents well-balanced per class accuracies, exhibiting notable improvements over PaCo.}
    \label{per_class_acc}
\end{figure}

Sometimes, the standard protocol of reporting many-shot (more than 100 images), medium-shot (20-100 images), and few-shot (less than 20 images) accuracies may not be sufficient to fully represent the effectiveness of long-tailed methods, especially when some extreme tail classes exhibit significantly low accuracies. In Fig.~\ref{per_class_acc}, we present detailed per-class accuracies of our method in three datasets. We compare our method with the widely recognized PaCo method on ImageNet-LT dataset, as only ImageNet-LT pre-trained weights are available for comparison. 

\subsection{Impact of initial word choice for Textual Inversion}
\begin{figure}[ht!]
    \centering
    \includegraphics[width=0.95\linewidth]{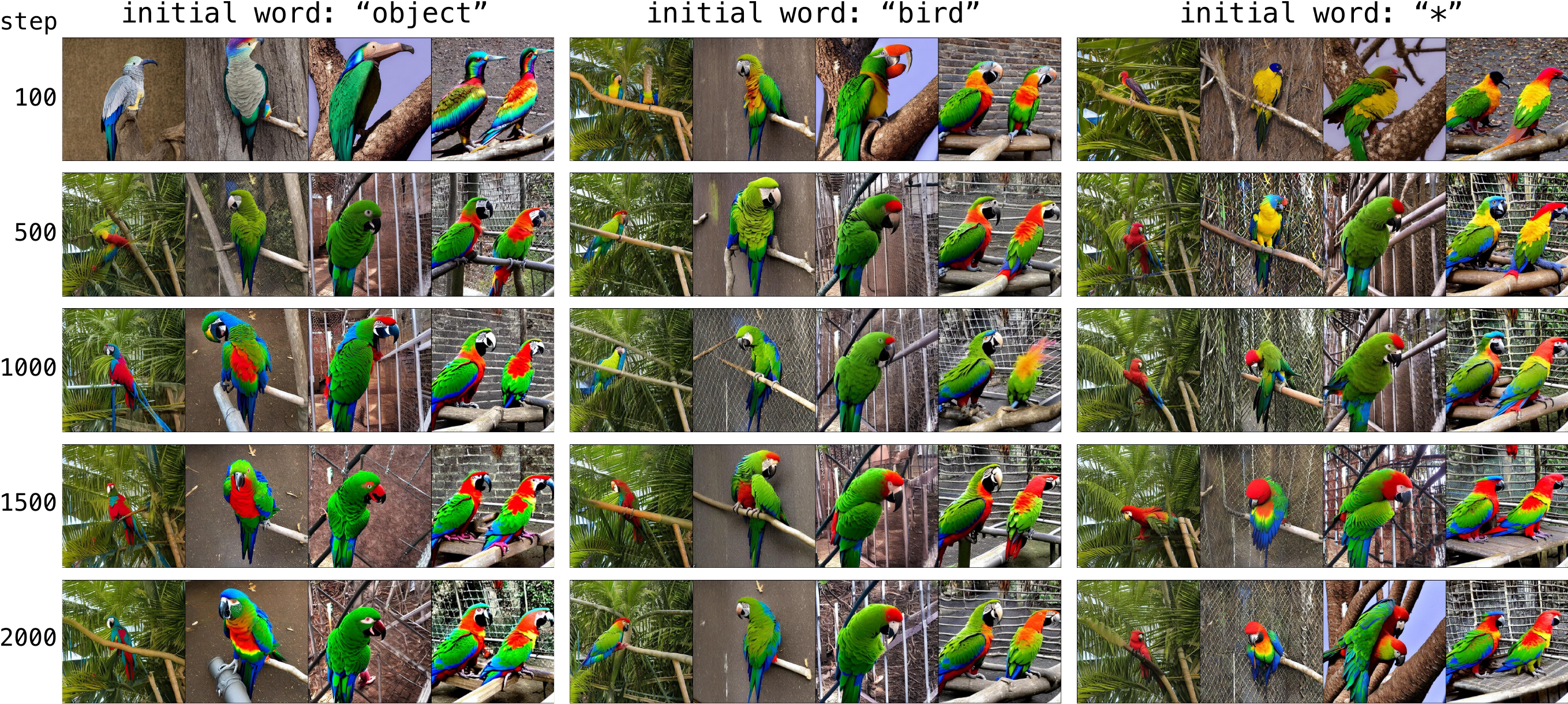}
    \caption{Optimization process of Textual Inversion with varying initialization of the token "*". We demonstrate the Textual Inversion training on six images from the macaw class of IN100-LT, displaying the intermediate images at each step of the optimization. We note that the optimization process of Textual Inversion is robust to the choice of initial words.}
    \label{init_word}
\end{figure}

One major advantage of using Textual Inversion is that prior knowledge related to the classes is not required. In Fig.~\ref{init_word}, we illustrate the optimization process of Textual Inversion with different initialization words. While it is true that choosing an initialization word closer to the concept of the actual class can accelerate convergence, we also observe reasonably fast and accurate convergence with more general initialization words such as "object" and "*". Hence, we initialize the token "*" with the CLIP embedding of "object" and continue optimization. Our method can effectively capture the image distribution without any prior knowledge from the text domain by solely relying on the image domain during the optimization process.

\subsection{Impact of varying diffusion parameters}

While our main focus is on discussing more complex generation strategies, it is important to note that changing basic parameters of the diffusion process, such as scale, can have a dramatic impact on the results. Our findings mostly align with the previous research by Sariyildiz~\etal~\cite{sariyildiz2023fake}. As suggested in Fig.~\ref{diffusion_steps}, the number of reverse process steps does not introduce significant performance gains when exceeding the default step of 50. Thus, regardless of the generation strategy, we keep our generation rule to 50 diffusion steps. We further explored adopting random steps between 0 and 100, but this didn't bring significant changes. 

As discussed earlier in diffusion communities, we observe that it is very important to fix at least one dimension of generated images' resolution to 512. Empirically, we notice that excessive scalings or scalings that do not maintain at least one dimension at 512 results in a notable degradation in image quality. Therefore, in our work, we adhere to the default image resolution of Stable Diffusion, which is 512$\times$512.

\subsection{Impact of the different prefix for single template}
Throughout our paper, we employ a single template (\textit{i.e.} "a photo of a \{\texttt{CLS}\}") as a strong baseline since it surprisingly outperforms most other methods, especially in lower guidance scales. Previous research by Sariyildiz~\etal~\cite{sariyildiz2023fake} suggests that the use of a simpler prompt, specifically "\{\texttt{CLS}\}", resulted in higher performance. However, due to the variations in experimental setups, particularly regarding the guidance scale and the version of Stable Diffusion used, we actually observed better results when using the prompt "a photo of a \{\texttt{CLS}\}". This prompt achieved a score of 59.9\%, whereas "\{\texttt{CLS}\}" achieved a score of 56.9\% in the scale 1.0 setting.

\subsection{Performance difference between Stable Diffusion v1 and v2}
As noted in the main paper, our experiments primarily utilize Stable Diffusion v1.5. Using a single template prompting strategy, we provide a comparison between v1 and v2 on different guidance scales in Table~\ref{sd2}. Following previous experiments, IN100 with 130K generated images were used. We adopt Stable Diffusion v2.1 base model in 512$\times$512 resolution for a fair comparison.
\begin{table}[h!]
\centering
\begin{minipage}[t]{0.45\linewidth}
\caption{Comparison of top-1 accuracies (\%) between Stable Diffusion v1 and v2 on IN100 with single template strategy ("a photo of a \{\texttt{CLS}\}"). It is important to note that scale 7.5 and 9.0 are the basic guidance scales of Stable Diffusion v1 and v2, respectively.}
\label{sd2}
\vspace{1mm}
\centering
\resizebox{1.0\textwidth}{!}{
\begin{tabular}{lccc}
    \cmidrule[1.0pt]{1-3}
    & \multicolumn{2}{c}{\textbf{IN100 Top-1 Accuracy~($\%$)}} \\ 
    \cmidrule[1.0pt]{2-3}
    \text{Scale} & \text{Stable Diffusion v1} & \text{Stable Diffusion v2} \\ 
    \cmidrule[1.0pt]{1-3}
    1.0 & 59.9 & 59.2 \\
    7.5 & 41.5 & 33.8 \\ 
    9.0 & 37.5 & 30.9 \\
    \cmidrule[1.0pt]{1-3}
\end{tabular}
}
\end{minipage}
\quad
\begin{minipage}[t]{0.45\linewidth}
    \centering
    \captionof{figure}{Impact of diffusion steps on top-1 accuracy.}
    \vspace{2mm}
    \includegraphics[width=0.9\linewidth]{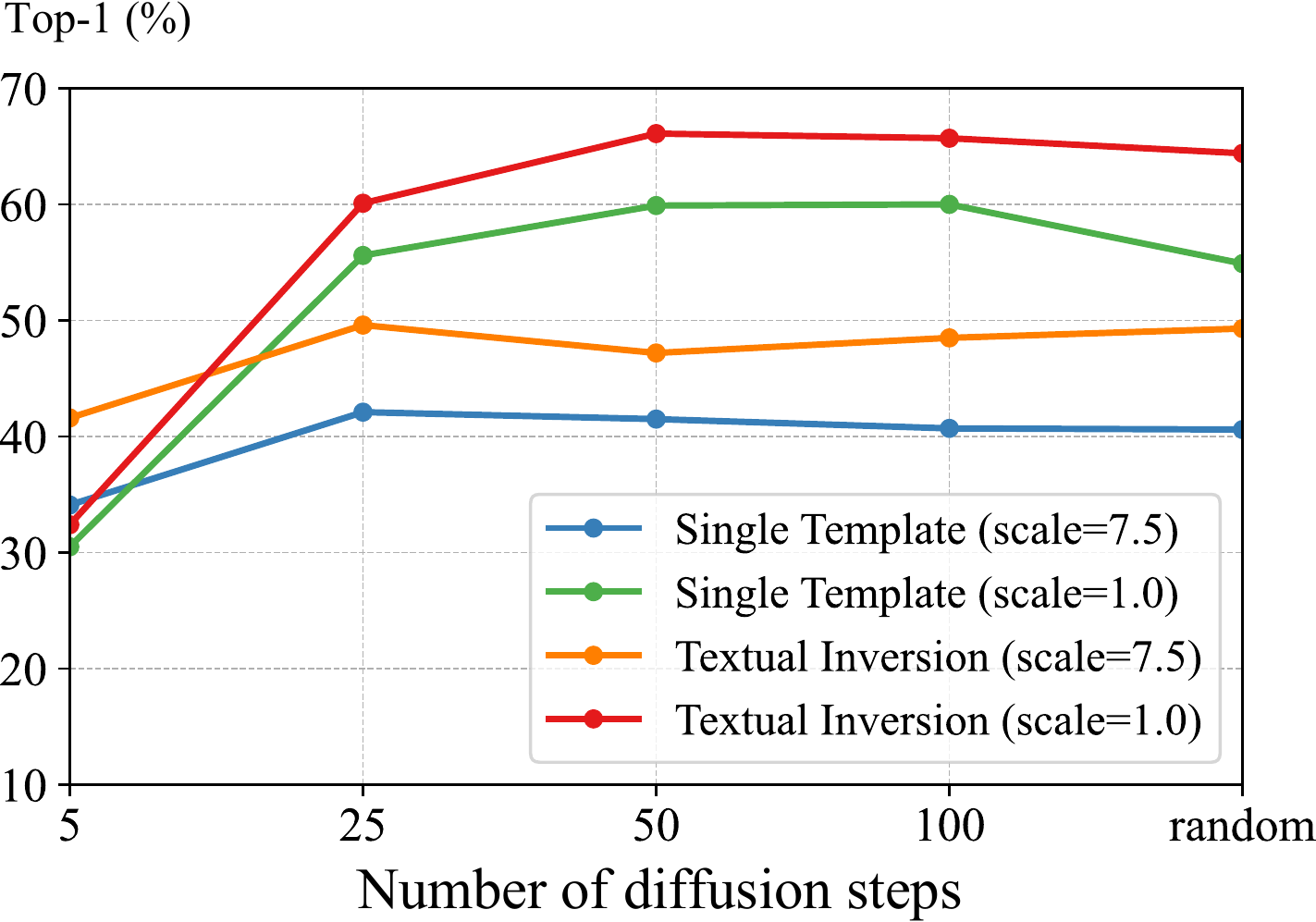}
    \label{diffusion_steps}
\end{minipage}
\end{table}
\section{Qualitative results}

\subsection{Generated prompts}
\label{sup_qualitative_generated_prompts}
We first present generated prompts using the methods mentioned above. Specifically, prompts for CLIP templates (Fig.~\ref{clip_template_prompt}), T5 (Fig.~\ref{t5_prompt}), Flan-T5 XXL (Fig.~\ref{flan-t5_prompt}), Prompt-Extend (Fig.~\ref{prompt-extend_prompt}), and the caption generated by BLIP2 (Fig.~\ref{caption_prompt}) are provided. For language models, we randomly select prompts for green iguana and African bush elephant classes. 

\begin{figure}[t]
\centering\noindent
\fbox{%
  \begin{minipage}[t]{0.5\textwidth}
    \parbox{\linewidth}{
        a bad photo of a \{\texttt{CLS}\}. \\
        a photo of many \{\texttt{CLS}\}. \\
        a sculpture of a \{\texttt{CLS}\}. \\
        a photo of the hard to see \{\texttt{CLS}\}. \\
        a low resolution photo of the \{\texttt{CLS}\}. \\
        a rendering of a \{\texttt{CLS}\}. \\
        graffiti of a \{\texttt{CLS}\}. \\
        a bad photo of the \{\texttt{CLS}\}. \\
        a cropped photo of the \{\texttt{CLS}\}. \\
        a tattoo of a \{\texttt{CLS}\}. \\
        the embroidered \{\texttt{CLS}\}. \\
        a photo of a hard to see \{\texttt{CLS}\}. \\
        a bright photo of a \{\texttt{CLS}\}. \\
        a photo of a clean \{\texttt{CLS}\}. \\
        a photo of a dirty \{\texttt{CLS}\}. \\
        a dark photo of the \{\texttt{CLS}\}. \\
        a drawing of a \{\texttt{CLS}\}. \\
        a photo of my \{\texttt{CLS}\}. \\
        the plastic \{\texttt{CLS}\}. \\
        a photo of the cool \{\texttt{CLS}\}. \\
        a close-up photo of a \{\texttt{CLS}\}. \\
        a black and white photo of the \{\texttt{CLS}\}. \\
        a painting of the \{\texttt{CLS}\}. \\
        a painting of a \{\texttt{CLS}\}. \\
        a pixelated photo of the \{\texttt{CLS}\}. \\
        a sculpture of the \{\texttt{CLS}\}. \\
        a bright photo of the \{\texttt{CLS}\}. \\    
        a cartoon \{\texttt{CLS}\}. \\
        art of a \{\texttt{CLS}\}. \\
        a sketch of the \{\texttt{CLS}\}. \\
        a embroidered \{\texttt{CLS}\}. \\
        a pixelated photo of a \{\texttt{CLS}\}. \\
        itap of the \{\texttt{CLS}\}. \\
        a jpeg corrupted photo of the \{\texttt{CLS}\}. \\
        a good photo of a \{\texttt{CLS}\}. \\
        a plushie \{\texttt{CLS}\}. \\
        a photo of the nice \{\texttt{CLS}\}. \\
        a photo of the small \{\texttt{CLS}\}. \\
        a photo of the weird \{\texttt{CLS}\}. \\
        the cartoon \{\texttt{CLS}\}.
    }
  \end{minipage}%
  \hfill
  \begin{minipage}[t]{0.5\textwidth}
    \parbox{\linewidth}{ 
        a cropped photo of a \{\texttt{CLS}\}. \\
        a plastic \{\texttt{CLS}\}. \\
        a photo of the dirty \{\texttt{CLS}\}. \\
        a jpeg corrupted photo of a \{\texttt{CLS}\}. \\
        a blurry photo of the \{\texttt{CLS}\}. \\
        a photo of the \{\texttt{CLS}\}. \\
        a good photo of the \{\texttt{CLS}\}. \\
        a rendering of the \{\texttt{CLS}\}. \\
        a \{\texttt{CLS}\} in a video game. \\
        a photo of one \{\texttt{CLS}\}. \\
        a doodle of a \{\texttt{CLS}\}. \\
        a close-up photo of the \{\texttt{CLS}\}. \\
        a photo of a \{\texttt{CLS}\}. \\
        the origami \{\texttt{CLS}\}. \\
        the \{\texttt{CLS}\} in a video game. \\
        a sketch of a \{\texttt{CLS}\}. \\
        a doodle of the \{\texttt{CLS}\}. \\
        a origami \{\texttt{CLS}\}. \\
        a low resolution photo of a \{\texttt{CLS}\}. \\
        the toy \{\texttt{CLS}\}. \\
        a rendition of the \{\texttt{CLS}\}. \\
        a photo of the clean \{\texttt{CLS}\}. \\
        a photo of a large \{\texttt{CLS}\}. \\
        a rendition of a \{\texttt{CLS}\}. \\
        a photo of a nice \{\texttt{CLS}\}. \\
        a photo of a weird \{\texttt{CLS}\}. \\
        a blurry photo of a \{\texttt{CLS}\}. \\    
        art of the \{\texttt{CLS}\}. \\
        a drawing of the \{\texttt{CLS}\}. \\
        a photo of the large \{\texttt{CLS}\}. \\
        a black and white photo of a \{\texttt{CLS}\}. \\
        the plushie \{\texttt{CLS}\}. \\
        a dark photo of a \{\texttt{CLS}\}. \\
        itap of a \{\texttt{CLS}\}. \\
        graffiti of the \{\texttt{CLS}\}. \\
        a toy \{\texttt{CLS}\}. \\
        itap of my \{\texttt{CLS}\}. \\
        a photo of a cool \{\texttt{CLS}\}. \\
        a photo of a small \{\texttt{CLS}\}. \\
        a tattoo of the \{\texttt{CLS}\}.
    }
  \end{minipage}%
}
\caption{A list of 80 templates used for CLIP templates strategy.}
\label{clip_template_prompt}
\end{figure}

\begin{figure}[ht]
\centering\noindent
\fbox{%
  \begin{minipage}[t]{0.5\textwidth}
    \parbox{\linewidth}{
    green iguana in a tropical rainforest \\ 
    iguana in green on the beach \\ 
    green iguana in the forest \\ 
    iguana in green on the beach \\ 
    green iguana in the field \\ 
    iguanas and a red and green iguana. \\ 
    iguana reversing the greens with a roar \\ 
    iguana in green on a sunny day \\ 
    iguana in green with a shady twig \\ 
    green iguana in a pond \\ 
    a lone iguana looks like it is a green. \\ 
    green iguana in a lagoon \\ 
    iguana in green \\ 
    green iguana in the wild. \\ 
    iguanas in the green of the bushes \\ 
    a lone iguana in green \\ 
    iguana in green on the green \\ 
    iguana in green on a sunny day 
    }
  \end{minipage}%
  \hfill
  \begin{minipage}[t]{0.5\textwidth}
    \parbox{\linewidth}{
        African elephants are grazing in the bush.\\
        elephants in the bush in africa. \\
        elephants in a bush in africa. \\
        an elephant in the bush in africa. \\
        A group of elephants in a bush in Africa. \\
        an elephant is grazing in a bush in africa. \\
        elephants in a bush in African. \\
        elephants and bush in the africa. \\
        A group of elephants in the bush in Africa. \\
        African elephants are in a bush. \\
        African elephants roaming in a bush. \\
        An elephant in the bush in Africa. \\
        An elephant in a bush in Africa. \\
        elephants in the bush in africa. \\
        elephants and buffaloes in the bush in africa. \\
        An elephant is in the bush in Africa. \\
        An elephant in a bush in Africa. \\
        An elephant in a bush in the country.
    }
  \end{minipage}%
}
\caption{Randomly selected prompts sampled by T5 language model. We display prompts for green iguana and African bush elephant classes, decoded using top-p nucleus sampling strategy.}
\label{t5_prompt}
\end{figure}

\begin{figure}[ht]
\centering\noindent
\fbox{%
  \begin{minipage}[t]{0.5\textwidth}
    \parbox{\linewidth}{
        two iguanas are standing next to each other \\
        a lizard is sitting on a rock \\
        three iguanas are laying on the grass \\
        a green iguana is sitting on a branch \\
        an iguana is sitting on top of a tree branch \\
        a large iguana laying on a log in a zoo \\
        a close up of an iguana with its head turned \\
        the lizard is brown \\
        several iguanas are sitting on the ground \\
        a green iguana is standing on some steps \\
        a woman holding a lizard on a beach \\
        a lizard is sitting on top of a rock \\
        an iguana is sitting on top of a tree stump \\
        a lizard is walking on the side of a fence \\
        an iguana is sitting in the branches of a tree \\
        a lizard sitting on top of a log in the woods \\
        an iguana on a rock with its long tail \\
        a lizard sitting on a rock in the sun
    }
  \end{minipage}%
  \hfill
  \begin{minipage}[t]{0.5\textwidth}
    \parbox{\linewidth}{
        a group of elephants walking around a pond \\
        a large elephant eating hay from a large pile \\
        elephants are the largest land mammals \\
        a baby elephant is being held by a soldier \\
        an elephant walking on a dirt road \\
        an elephant standing in the grass \\
        an elephant is standing in tall grass \\
        two elephants walking in the grass \\
        an elephant standing in a field \\
        an elephant with tusks \\
        a herd of elephants standing in the water \\
        an elephant is walking through the brush \\
        a woman is petting an elephant \\
        a group of people riding on an elephant \\
        a baby elephant walking \\
        two elephants walking in the grass \\
        an elephant with a big nose \\
        two elephants standing in water 
    }
  \end{minipage}%
}
\caption{Randomly selected prompts captioned by BLIP2 from IN100-LT images. We display prompts for green iguana and African bush elephant classes.}
\label{caption_prompt}
\end{figure}

\begin{figure}[ht]
\centering\noindent
\fbox{%
\begin{minipage}[t]{1.0\textwidth}
    A green iguana rests on a tree branch in the forest.\\
    A green iguana curled up in the sun, its fur gleaming in the sunlight.\\
    A green iguana scurries through the lush forest in search of food.\\
    A green iguana is pacing across a lush tropical forest, its green fur gleaming in the sun.\\
    Imagine a green iguana sitting in a tree, sunbathing and enjoying the warm summer air.\\
    Imagine a green iguana standing in the middle of the tropical jungle.\\
    A green iguana scurries through the jungle, looking for a new home.\\
    The green iguana stood on a small island, its fur green and its eyes glowing in the sunlight.\\
    A green iguana is swimming in the ocean.
    \par\vspace{-1.0ex} 
    \par\rule{\linewidth}{0.4pt} 
    An African bush elephant grazing in the grasslands with its trunk up in the air.\\
    A herd of African bush elephants slowly strolling through the vast plains.\\
    The African bush elephant trudged through the tall grass, its trunk in the air.\\
    A majestic African bush elephant strolls through a lush, green forest.\\
    The African bush elephants walked peacefully across the lush green grasslands.\\
    A majestic African bush elephant gracefully walking through the dense foliage.\\
    Imagine a majestic African bush elephant grazing in the lush green jungle.\\
    An African bush elephant saunters through the forest, its large ears floppy in the heat.\\
    Imagine an African bush elephant galloping through the forest.
\end{minipage}%
}
\caption{Randomly selected prompts sampled by Flan-T5 XXL language model. We display prompts for green iguana and African bush elephant classes, decoded using top-p nucleus sampling strategy.}
\label{flan-t5_prompt}
\end{figure}

\begin{figure}[ht]
\centering\noindent
\fbox{%
\begin{minipage}{1.0\textwidth}
    green iguana with a red mohawk, green mohawk, wearing sunglasses, synthwave style, \\
    green iguana with a white nose and a green frog face, photograph, national geographic, \\
    green iguana, beautiful landscape, environment, colors, film, dramatic, cinematic, \\ 
    green iguana, intricate, elegant, highly detailed, digital painting, artstation, \\
    green iguana, wildlife photography  on a road, high definition, shot with sigma f/ 4.2, \\
    green iguana with a flower face  on its head, natural light, sharp, detailed face, \\
    green iguana  in jungle,  Beksinski painting, part by Adrian Ghenie  and Gerhard Richter. \\
    green iguana with a gun in its mouth, realistic, digital art, painted by seb mckinnon,\\
    green iguana hybrid  in the style of pixar, cinematic composition, cinematic lighting, 
    \par\vspace{-1.0ex} 
    \par\rule{\linewidth}{0.4pt} 
    African bush elephant portrait, face, wearing a crown, futuristic, \\
    African bush elephant as a fantasy D\&D character, portrait art by Donato Giancola and, \\
    African bush elephant with long hairs in fur and fur and fluffy fur with pastel colors, \\
    African bush elephant robot, 3d model, photorealistic, 3d, rendered, hd, 4k, HD, sharp,\\
    African bush elephant hybrid with a peacock tail and rainbow mane, rainbow mane, pastel,\\
    African bush elephant with a big trunk and a trunk. A frog is sitting on top of a leaf. \\
    African bush elephant chimera, 4k, high detail, high-resolution photograph, professional,\\
    African bush elephant, award winning nature photography, National Geographic, \\
    African bush elephant chimera, 4k, DSLR, highly detailed, professional photography,
\end{minipage}%
}
\caption{Randomly selected prompts sampled by Prompt-Extend language model. We display prompts for green iguana and African bush elephant classes, decoded using top-p nucleus sampling strategy. We truncate lengthy sentences that consisted of a maximum of 60 tokens in order to fit the figure.}
\label{prompt-extend_prompt}
\end{figure}

\subsection{Generated images}
\label{sup_qualitative_generated_images}
\begin{figure}[t!]
    \centering
    \includegraphics[width=1.0\linewidth]{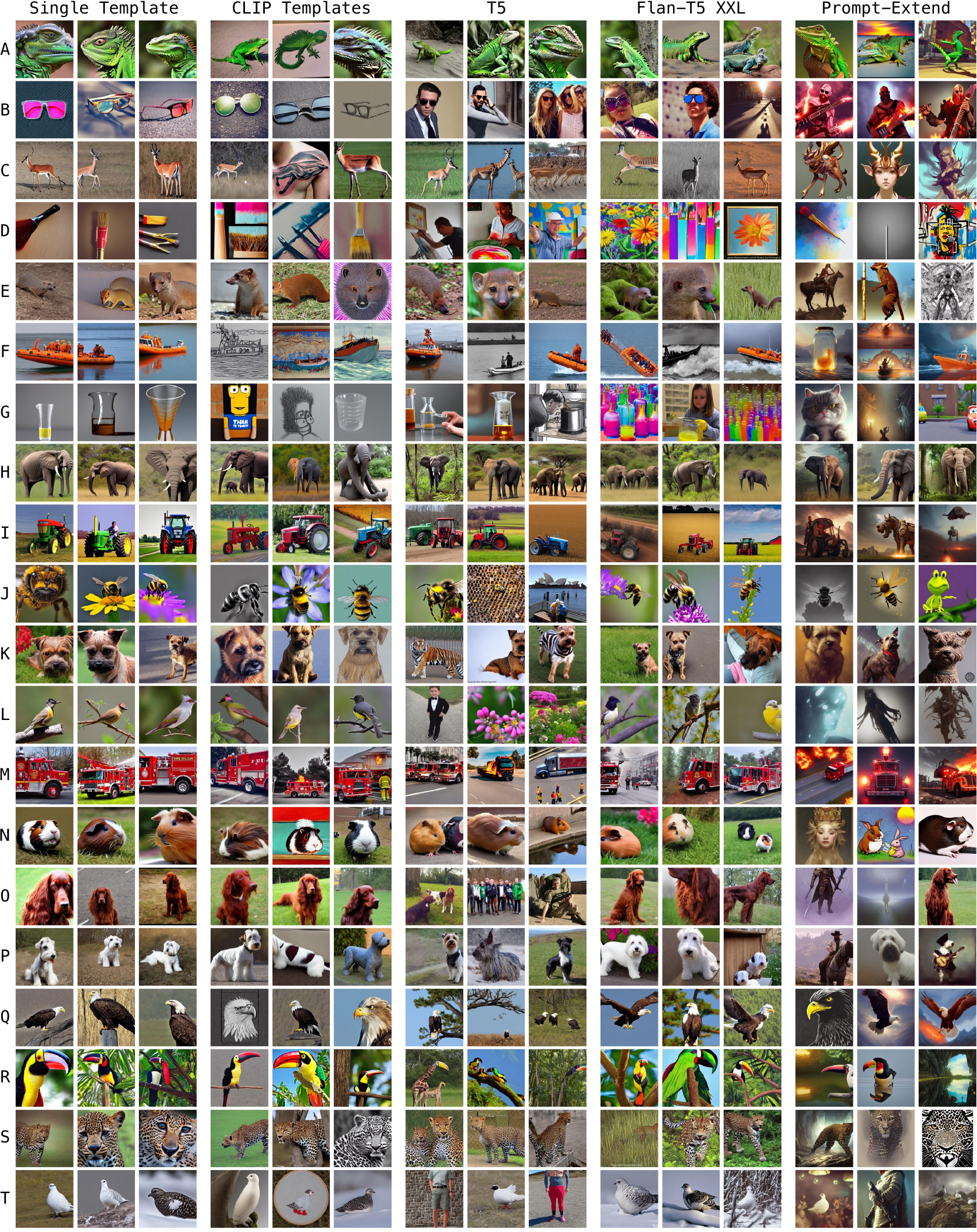}
    \caption{Randomly selected images from IN100 using different prompt-to-image based methods (scale=7.5). Class names are as follows:\{A: green iguana, B: sunglasses, C: gazelle, D: paintbrush, E: mongoose, F: lifeboat, G: beaker, H: African bush elephant, I: tractor, J: bee, K: Border Terrier, L: bulbul, M: firetruck, N: guinea pig, O: Irish Setter, P: Sealyham Terrier, Q: bald eagle, R: toucan, S: leopard, T: ptarmigan\}}
    \label{p2i_7.5_images}
\end{figure}

\begin{figure}[t!]
    \centering
    \includegraphics[width=1.0\linewidth]{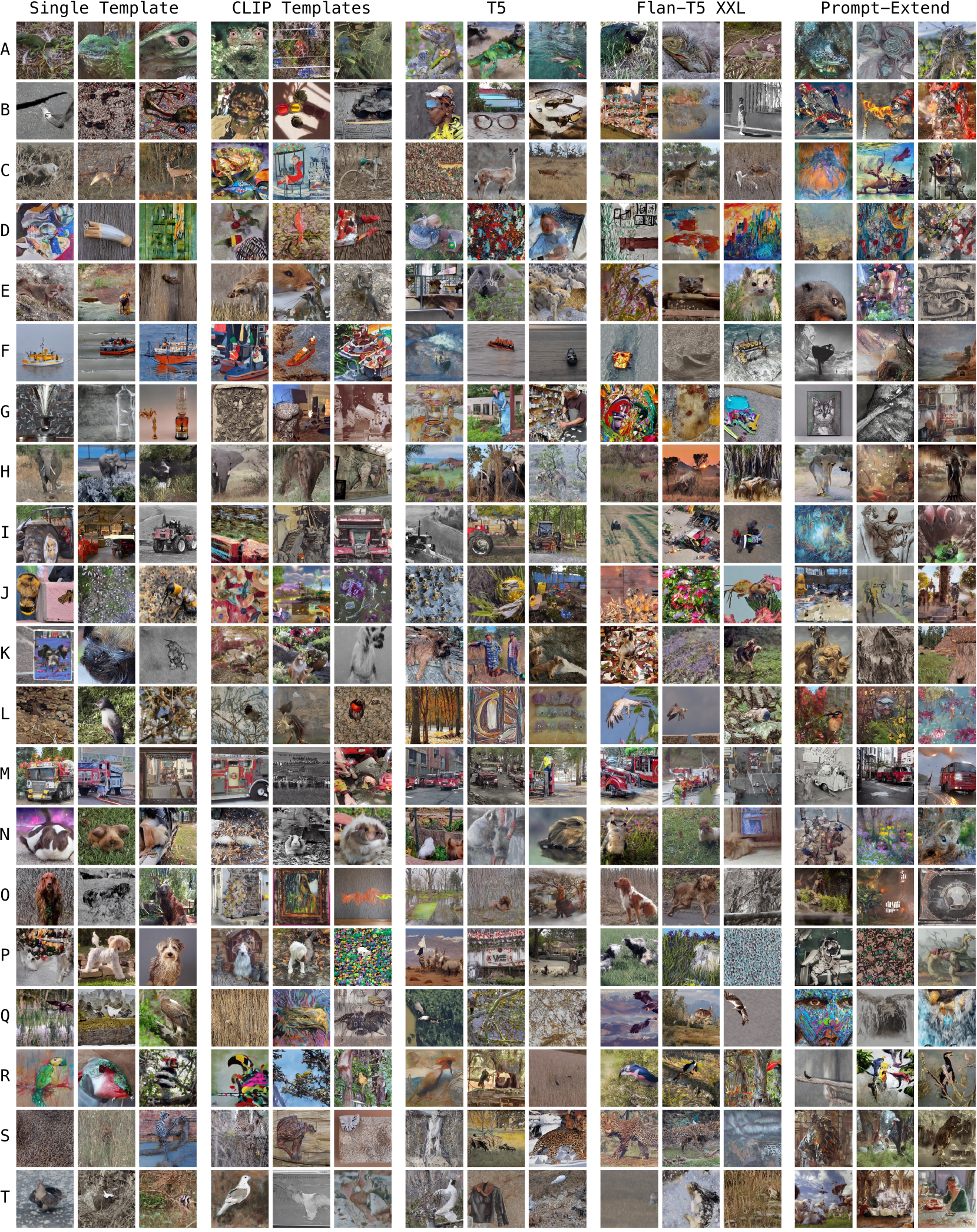}
    \caption{Randomly selected images from IN100 using different prompt-to-image based methods (scale=1.0). Class names are as follows:\{A: green iguana, B: sunglasses, C: gazelle, D: paintbrush, E: mongoose, F: lifeboat, G: beaker, H: African bush elephant, I: tractor, J: bee, K: Border Terrier, L: bulbul, M: firetruck, N: guinea pig, O: Irish Setter, P: Sealyham Terrier, Q: bald eagle, R: toucan, S: leopard, T: ptarmigan\}}
    \label{p2i_1.0_images}
\end{figure}

\begin{figure}[t!]
    \centering
    \includegraphics[width=1.0\linewidth]{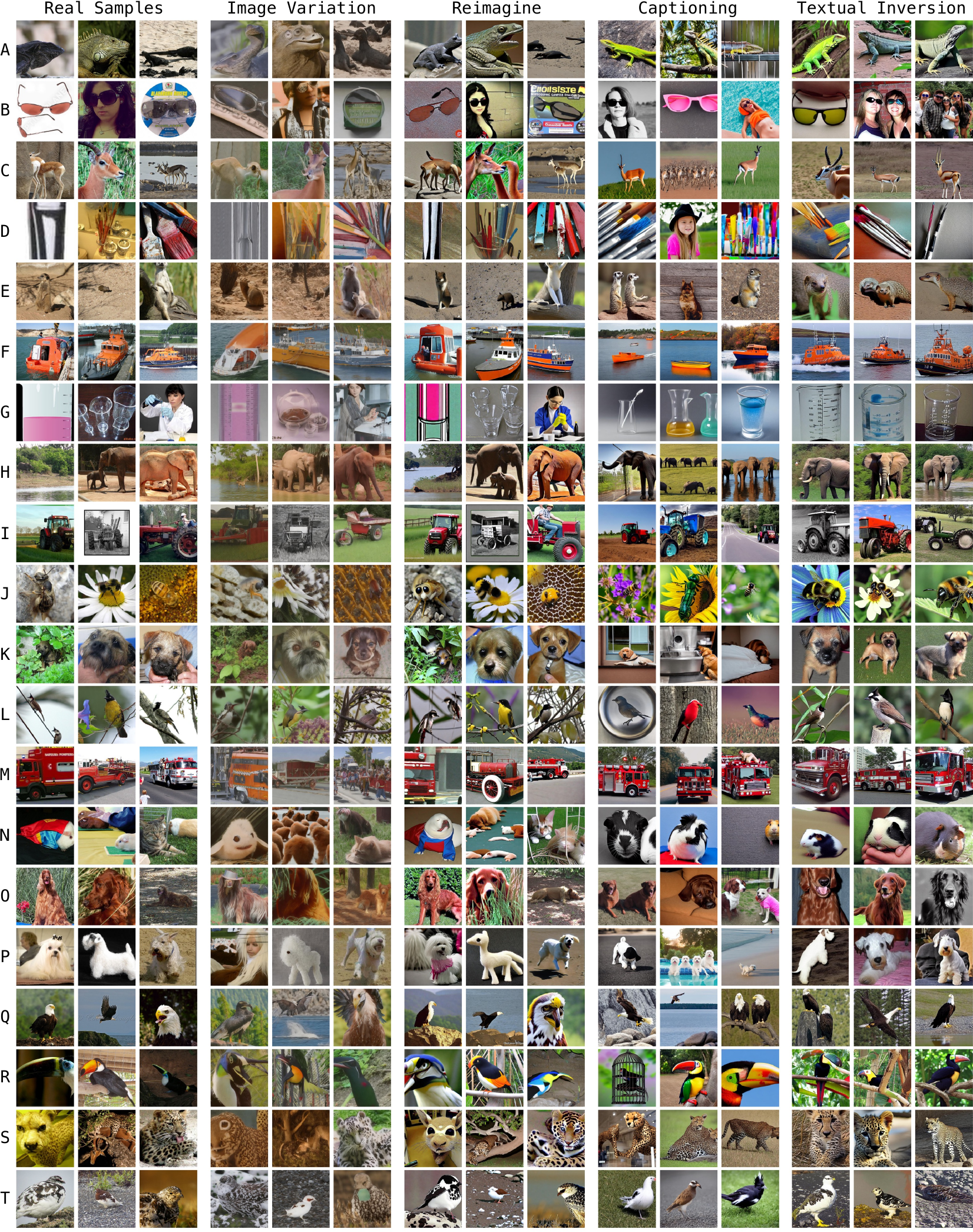}
    \caption{Randomly selected images from IN100 using image-to-image and transmodal based methods (scale=7.5). IN100-LT samples were used as source images. Class names and the number of real samples are as follows:\{A: (green iguana, 1000), B: (sunglasses, 765), C: (gazelle, 651), D: (paintbrush, 472), E: (mongoose, 342), F: (lifeboat, 223), G: (beaker, 180), H: (African bush elephant, 153), I: (tractor, 117), J: (bee, 80), K: (Border Terrier, 61), L: (bulbul, 55), M: (firetruck, 40), N: (guinea pig, 32), O: (Irish Setter, 26), P: (Sealyham Terrier, 21), Q: (bald eagle, 17), R: (toucan, 12), S: (leopard, 8), T: (ptarmigan, 5)\} }
    \label{rest_7.5_images}
\end{figure}

\begin{figure}[t!]
    \centering
    \includegraphics[width=1.0\linewidth]{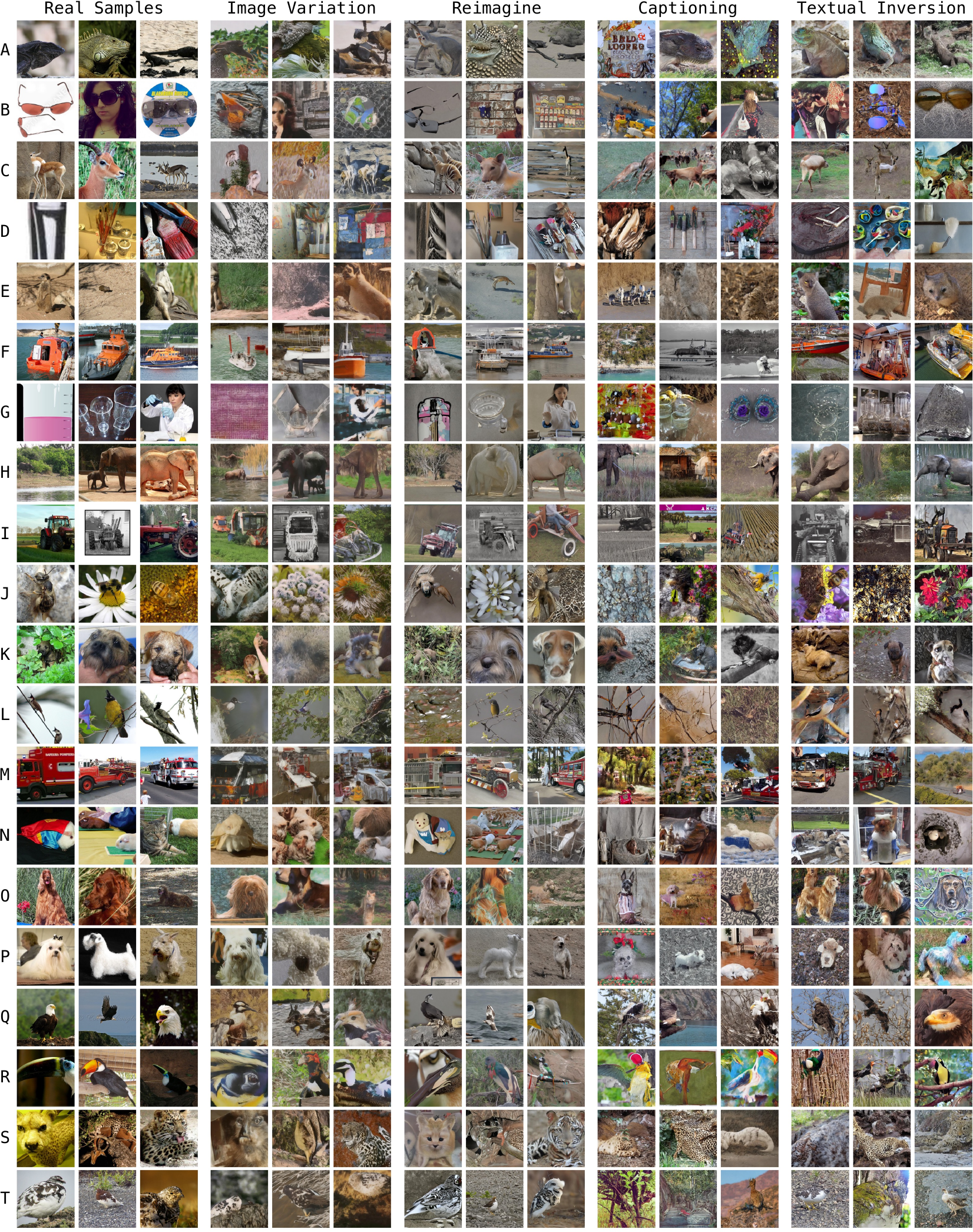}
    \caption{Randomly selected images from IN100 using image-to-image and transmodal based methods (scale=1.0). IN100-LT samples were used as source images. Class names and the number of real samples are as follows:\{A: (green iguana, 1000), B: (sunglasses, 765), C: (gazelle, 651), D: (paintbrush, 472), E: (mongoose, 342), F: (lifeboat, 223), G: (beaker, 180), H: (African bush elephant, 153), I: (tractor, 117), J: (bee, 80), K: (Border Terrier, 61), L: (bulbul, 55), M: (firetruck, 40), N: (guinea pig, 32), O: (Irish Setter, 26), P: (Sealyham Terrier, 21), Q: (bald eagle, 17), R: (toucan, 12), S: (leopard, 8), T: (ptarmigan, 5)\} }
    \label{rest_1.0_images}
\end{figure}

\begin{figure}[!ht]
    \centering
    \includegraphics[width=1.0\linewidth]{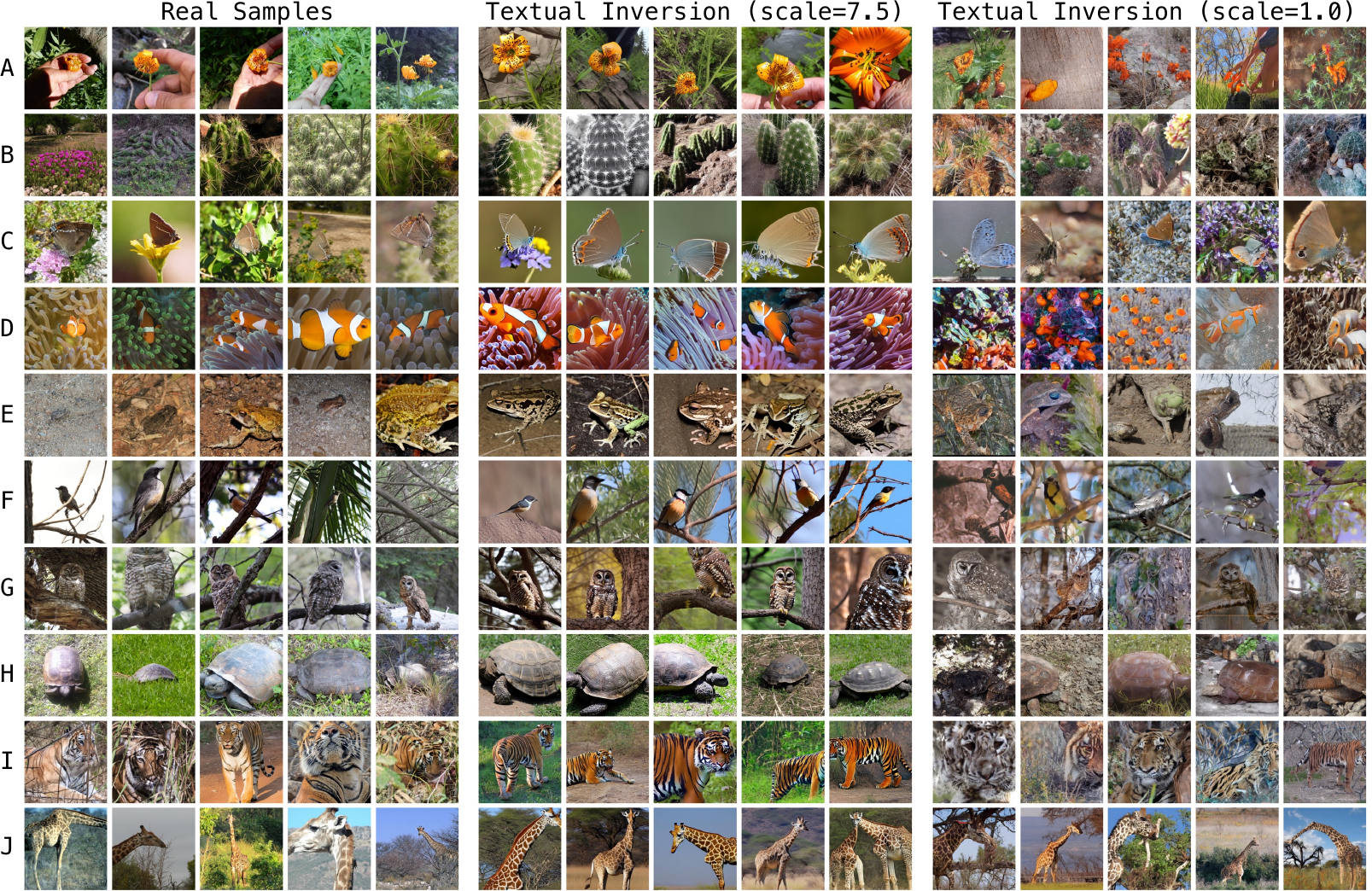}
    \caption{Qualitative results for iNaturalist2018. Due to the lack of class names, we list class indices. \{A: 5533, B: 6334, C: 1211, D: 2457, E: 2616, F: 3508, G: 3976, H: 4543, I: 4094, J: 4071\}}
    \label{iNaturalist-images}
\end{figure}

\begin{figure}[!ht]
    \centering
    \includegraphics[width=1.0\linewidth]{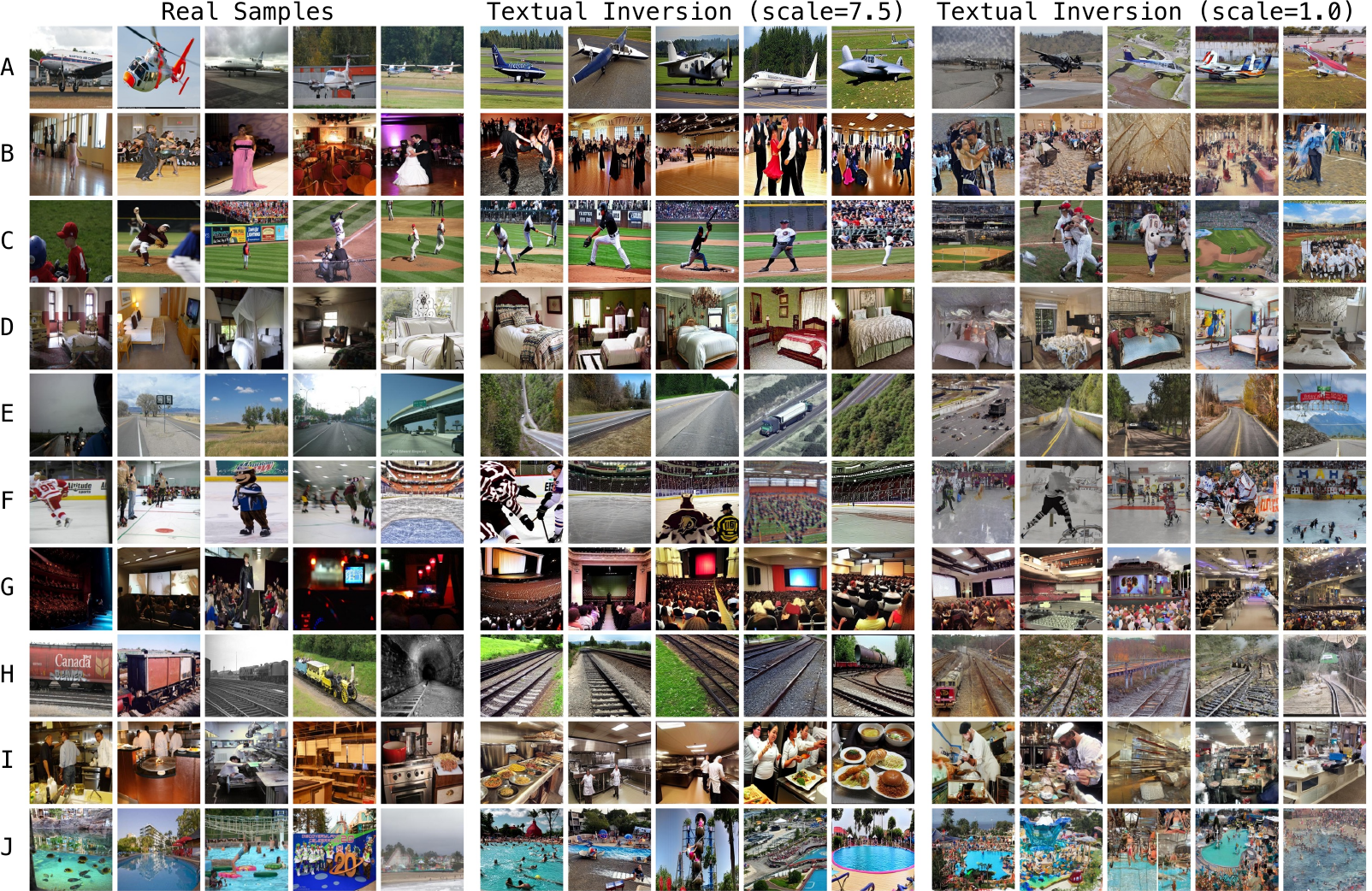}
    \caption{Qualitative results for Places-LT. Class names are as follows: \{A: airfield, B: ballroom, C: baseball field, D: bedroom, E: highway, F: ice skating rink indoor, G: movie theater indoor, H: railroad track, I: restaurant kitchen, J: water park\}}
    \label{places-lt_images}
\end{figure}

We present images generated by prompt-to-image based methods in scale 7.5 (Fig.~\ref{p2i_7.5_images}) and scale 1.0 (Fig.~\ref{p2i_1.0_images}), and image-to-image and transmodal methods in scale 7.5 (Fig.~\ref{rest_7.5_images}) and scale 1.0 (Fig.~\ref{rest_1.0_images}). We also provide samples for iNaturalist2018 in Fig.~\ref{iNaturalist-images}, and Places-LT in Fig.~\ref{places-lt_images}. Overall, images with higher guidance scale of 7.5 show better visual results when compared to those with lower guidance scale of 1.0. Furthermore, both the image-to-image and transmodal methods demonstrate improved alignment with real data. Especially, when utilizing Textual Inversion with a higher guidance scale, we observe the generation of extremely realistic images across various datasets. While lower guidance scale produce relatively noisier outputs, the notably higher top-1 accuracies obtained from lower guidance suggest the sparse existence of important samples. Although we rarely encounter samples where added noise actually results in plausible variations, we think that scaling these noisy samples, which yields more important samples over time, is the key reason for its high performance.

\clearpage
\section{Broader impacts and compute resources}
\label{sup_broader}
Fill-Up aims to mitigate the challenge of imbalanced real-world distributions by leveraging large-scale generative models. Given the brittle nature of training generative models under imbalanced data and the heavy computational resources required for full fine-tuning of state-of-the-art models, our approach, which focuses on lightweight optimization of a single text token per class, shows a promising direction for handling long-tailed distributions. In the domain of long-tailed recognition, where synthetic data generation approaches have relatively lagged behind state-of-the-art methods, our approach provides strong intuitions and promising results by leveraging a pre-trained generative model. With the rapid development of cutting-edge generative models, we expect  more approaches based on synthetic data generation to be actively discussed and explored in the field. 

Recent explosive developments in generative modeling brought forth numerous potential dangers, encompassing ethical concerns, privacy issues, legal implications, and much more. Built on top of recent advancements, our method also inherently shares potential risks and challenges associated with these models. Two specific issues that we have observed pertain to privacy and nudity, particularly in classes that involve human subjects. In specific classes such as sunscreen, where individuals wearing exposed clothing are prevalent, the optimization process of Textual Inversion may yield unintended images that could potentially be deemed NSFW (Not Suitable For Work). This issue can be further exacerbated by the tendency of generative models to memorize instances, such as real human beings, that they encounter during the training process. Safeguarding privacy and maintaining high ethical standards are our top priorities throughout this work. We will take additional precautions in the future application and release of any related assets. Our code and selected data will be made public upon paper publication after careful review.

In terms of computing resources, we employ a combination of RTX 3090 and A100 GPUs. With a single RTX 3090, Textual Inversion takes approximately 30 minutes per class, following our heuristic, on ImageNet-LT dataset. Additionally, sampling a single image with a scale of 1.0 takes about 3 seconds. For classification, training ResNet50 on ImageNet-LT with 2.6M filled-up images requires one day of computation with eight A100 GPUs. Computational requirements for IN100 ablation, Places-LT, and iNaturalist2018 scale with mentioned timings.

\end{document}